\documentclass[journal]{IEEEtran}
\usepackage{ifpdf}
\usepackage{cite}
\usepackage{amsmath,amssymb,amsfonts}
\usepackage{algorithmic}
\usepackage{graphicx}
\usepackage{textcomp}

\usepackage[caption=false,font=footnotesize]{subfig}
\usepackage{booktabs}
\usepackage{url}
\usepackage{caption}
\usepackage{amssymb}
\usepackage{acro}
\usepackage{multirow}
\usepackage{color}
\usepackage{subfiles} 

\DeclareAcronym{P3D}{short=P3D, long=pseudo 3D}
\DeclareAcronym{UB}{short=UB, long=upper bound}
\DeclareAcronym{FP}{short=FP, long=false positive}
\DeclareAcronym{PR}{short=PR, long=precision/recall}
\DeclareAcronym{AP}{short=AP, long=average precision}
\DeclareAcronym{AUC}{short=AUC, long=area under curve}
\DeclareAcronym{CT}{short=CT, long=computed tomography}
\DeclareAcronym{DGL}{short=DGL, long=deep growing learning}
\DeclareAcronym{MAP}{short=MAP, long=mean average precision}
\DeclareAcronym{IoU}{short=IoU, long=intersection over union}
\DeclareAcronym{3DCCN}{short=CECN, long=contextual-enhanced CenterNet}  
\DeclareAcronym{FROC}{short=FROC, long=free-response ROC curve}
\DeclareAcronym{LPG}{short=LPG, long=lesion proposal generator}
\DeclareAcronym{LPC}{short=LPC, long=lesion proposal classifier}
\DeclareAcronym{R@95P}{short=R@95P, long=recall at 95\% precision}
\DeclareAcronym{R@90P}{short=R@90P, long=recall at 90\% precision}
\DeclareAcronym{R@85P}{short=R@85P, long=recall at 85\% precision}
\DeclareAcronym{R@80P}{short=R@80P, long=recall at 80\% precision}
\DeclareAcronym{CNN}{short=CNN, long=convolutional neural network, foreign-plural={}}
\DeclareAcronym{HNC}{short=OBHS, long=overlap-based hard sampling}
\DeclareAcronym{OBSS}{short=OBSS, long=overlap-based soft sampling}
\DeclareAcronym{MGTM}{short=MGTM, long=missing ground-truth mining}
\DeclareAcronym{OBHS}{short=OBHS, long=overlap-based hard sampling}
\DeclareAcronym{SLLP}{short=SLLP, long=slice-level label propagation}
\DeclareAcronym{HNSL}{short=HNSL, long=hard negative suppression loss}
\DeclareAcronym{GLC}{short=GLC, long=global-local classifier}
\DeclareAcronym{MVCNN}{short=GLC-MV, long=global-local classifier with multi-view input}
\DeclareAcronym{GL3D}{short=GLC-3D, long=global-local classifier with 3D input}
\DeclareAcronym{MLC3D}{short=MLC3D, long=multi-level contextual 3-D}
\DeclareAcronym{MULAN}{short=MULAN, long=multitask universal lesion analysis network}
\DeclareAcronym{RECIST}{short=RECIST, long=response evaluation criteria in solid tumours}
\DeclareAcronym{PACS}{short=PACS, long=picture archiving and communication system, foreign-plural={}}

\newcommand{\Sec}{Sec.}
\newcommand{\Fig}{Fig.}
\newcommand{\Eq}{Eq.}
\newcommand{\Tab}{Table}

\def\onedot{.}
\def\eg{\emph{e.g}\onedot} 
\def\ie{\emph{i.e}\onedot}

\def\etal{\emph{et al}\onedot}

\def\BibTeX{{\rm B\kern-.05em{\sc i\kern-.025em b}\kern-.08em
    T\kern-.1667em\lower.7ex\hbox{E}\kern-.125emX}}
\begin{document}
\title{Lesion-Harvester: Iteratively Mining Unlabeled Lesions and Hard-Negative Examples at Scale}
\author{Jinzheng Cai, Adam P. Harrison, Youjing Zheng, Ke Yan, Yuankai Huo, Jing Xiao, Lin Yang, Le Lu
\thanks{J. Cai, A. P. Harrison, K. Yan, Y. Huo, and L. Lu are with PAII Inc., Bethesda, MD, USA. Y. Huo is now with Vanderbilt University, Nashville, TN, USA. (e-mail: caijinzheng883@paii-labs.com, adampharrison070@paii-labs.com, yanke383@paii-labs.com)}
\thanks{Y. Zheng is with Virginia Polytechnic Institute and State University, Blacksburg, VA, USA. (email: zhengyoujing@vt.edu)}
\thanks{J. Xiao is with Ping An Insurance (Group) Company of China, Ltd., Shenzhen, PRC.}
\thanks{L. Yang is with University of Florida, Gainesville, FL, USA.}
\thanks{Corresponding author: Jinzheng Cai}
\thanks{Preprint, to appear in IEEE Transactions on Medical Imaging (https://doi.org/10.1109/TMI.2020.3022034).}
}

\maketitle

\begin{abstract}
 The acquisition of large-scale medical image data, necessary for training machine learning algorithms, is hampered by associated expert-driven annotation costs. Mining hospital archives can address this problem, but labels often incomplete or noisy, \eg{}, $\mathbf{50\%}$  of the lesions in DeepLesion are left unlabeled. Thus, effective label harvesting methods are critical. This is the goal of our work, where we introduce Lesion-Harvester---a powerful system to harvest missing annotations from lesion datasets at high precision. Accepting the need for some degree of expert labor, we use a small fully-labeled image subset to intelligently mine annotations from the remainder. To do this, we chain together a highly sensitive lesion proposal generator (LPG) and a very selective lesion proposal classifier (LPC). Using a new hard negative suppression loss, the resulting harvested and hard-negative proposals are then employed to iteratively finetune our LPG. While our framework is generic, we optimize our performance by proposing a new 3D contextual LPG and by using a global-local multi-view LPC. Experiments on DeepLesion demonstrate that Lesion-Harvester can discover an additional $\mathbf{9,805}$ lesions at a precision of $\mathbf{90\%}$. We publicly release the harvested lesions, along with a new test set of \textit{completely} annotated DeepLesion volumes. We also present a pseudo 3D IoU evaluation metric that corresponds much better to the real 3D IoU than current DeepLesion evaluation metrics. To quantify the downstream benefits of Lesion-Harvester we show that augmenting the DeepLesion annotations with our harvested lesions allows state-of-the-art detectors to boost their average precision by $7$ to $10\%$.
\end{abstract}
  
\begin{IEEEkeywords}
Lesion harvesting, lesion detection, hard negative mining, pseudo 3D IoU.
\end{IEEEkeywords}

%
\IEEEpeerreviewmaketitle

\section{Introduction}
\IEEEPARstart{P}{aralleling} developments in computer vision, recent years have seen the emergence of large-scale medical image databases~\cite{DBLP:journals/corr/abs-1712-06957,DBLP:conf/wacv/WangLSKBNYS17,DBLP:conf/cvpr/WangPLLBS17,yan_2018_deeplesion,DBLP:conf/cvpr/YanWLZHBS18}. These are seminal milestones in medical imaging analysis research that help address the data-hungry needs of deep learning and other machine learning technologies. Yet, most of these databases are collected retrospectively from hospital \acp{PACS}, which house the medical image and text reports from daily radiological workflows. While harvesting \acp{PACS} will likely be essential toward truly obtaining large-scale medical imaging data~\cite{DBLP:journals/jdi/KohliSG17}, their data are entirely ill-suited for training machine learning systems~\cite{harvey2019standardised} as they are not curated from a machine learning perspective. As a result, popular large-scale medical imaging datasets suffer from  uncertainties, mislabelings~\cite{DBLP:conf/cvpr/WangPLLBS17,DBLP:conf/aaai/IrvinRKYCCMHBSS19,DBLP:conf/micad/CalliSSMG19} and incomplete annotations~\cite{DBLP:conf/cvpr/YanWLZHBS18}, a trend that promises to increase as more and more \ac{PACS} data is exploited. Correspondingly, there is a great need for effective data curation, but, unlike in computer vision, these problems cannot be addressed by crowd-sourcing approaches~\cite{DBLP:conf/cvpr/DengDSLL009,DBLP:conf/eccv/LinMBHPRDZ14}. Instead this need calls for alternative methods tailored to the demanding medical image domain. This is the focus of our work, where we articulate a powerful and effective label completion framework for lesion datasets, applying it to harvest unlabeled lesions from the recent DeepLesion dataset~\cite{DBLP:conf/cvpr/YanWLZHBS18}. 

\begin{figure}[t!]
    \centering
    \includegraphics[trim=0in 4in 4in 0in,clip,width=\linewidth]{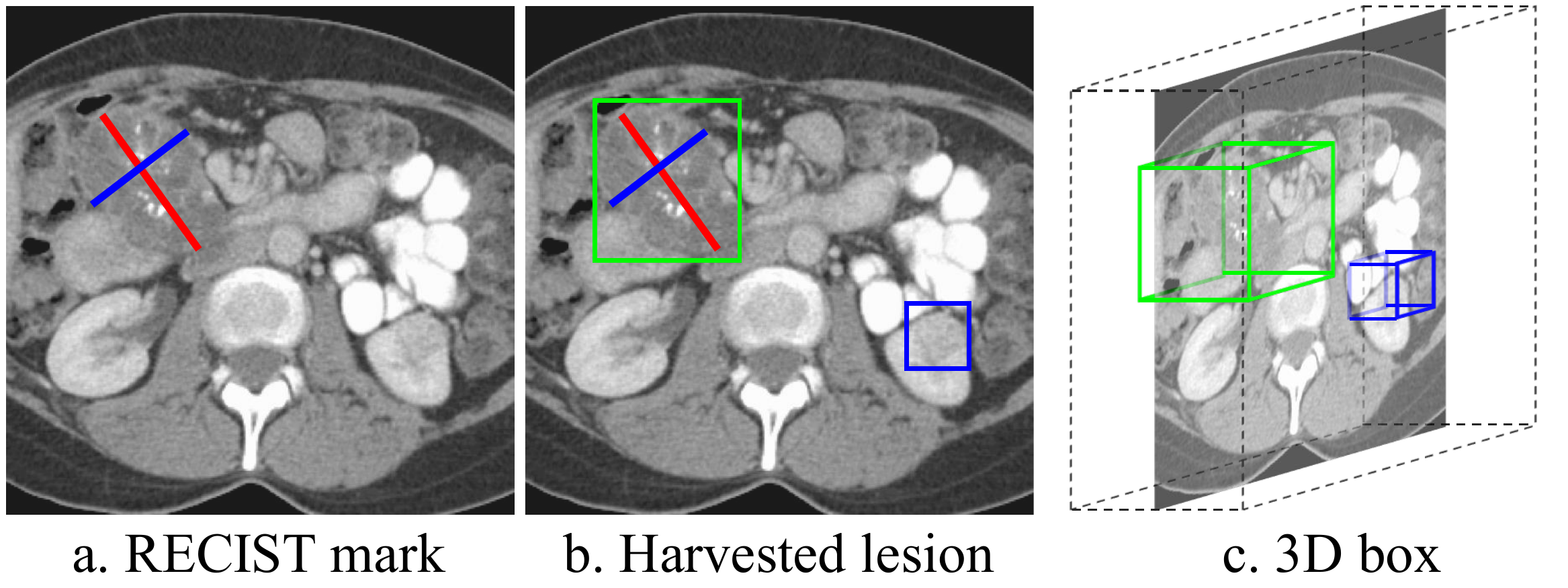}
    \caption{(a) depicts an example of a \ac{RECIST} marked CT slice. \ac{RECIST} marks may be incomplete by both not including co-exsiting lesions, \eg{},~the blue 2D box in (b), or by not covering the 3D extent of lesions in other slices, \eg{}, the green and blue \emph{3D} boxes in (c). We aim to complete lesion annotations in both senses of (b) and (c).}
    \label{fig:0}
\end{figure}

DeepLesion~\cite{yan_2018_deeplesion,DBLP:conf/cvpr/YanWLZHBS18} is a recent publicly released medical image database of CT sub-volumes along with localizations of lesions. These were mined from \ac{CT} scans from the US National Institutes of Health Clinical Center \ac{PACS}. The mined lesions were extracted from \ac{RECIST}~\cite{Eisenhauer2009} marks performed by clinicians to measure tumors in their daily workflow. See \Fig~\ref{fig:0}(a) for an example of a \ac{RECIST} marked lesion. 
In total, DeepLesion contains $32,735$ retrospectively clinically annotated lesions from $10,594$ \ac{CT} scans of $4,427$ unique patients. A variety of lesion types and subtypes have been included in this database, such as lung nodules, liver tumors, and enlarged lymph nodes. As such, the DeepLesion dataset is an important source of data for medical imaging analysis tasks, including training and characterizing lesion detectors and for developing radiomics-based biomarkers for tumor assessment and tracking. However, due to the \ac{RECIST} guidelines~\cite{Eisenhauer2009} and workload limits, physicians typically marked only a small amount of lesions per \ac{CT} scan as the \emph{finding(s) of interest}. Yet, as shown in \Fig~\ref{fig:0}(b), more often than not \ac{CT} images exhibit multiple co-existing lesions per patient. Indeed, based on a recent empirical study~\cite{ieee/miccai/YanBS18}, and our own results presented later, there are about the same quantity of missing findings compared to reported ones. Moreover, as \Fig~\ref{fig:0}(c) illustrates, \ac{RECIST} marks do not indicate the 3D extent, leaving tumor regions of the same instance in adjoining slices unmarked. This severely challenges the development of high-fidelity disease detection algorithms and artificially limits the dataset's usefulness for biomarker development. Nevertheless, it is highly impractical and infeasible to recruit physicians to manually revise and add back annotations for the entire database.

To address this issue, we aim to reliably discover and harvest unlabeled lesions. Given the expert-driven nature of annotations, our approach, named Lesion-Harvester, accepts the need for a small amount of supplementary physician labor. It integrates three processes: (1) a highly sensitive detection-based \ac{LPG} to generate lesion candidates, (2) manual verification of a small amount of the lesion proposals, and (3) a very selective \ac{LPC} that uses the verified proposals to automatically harvest prospective positives and hard negatives from the rest. These processes are tied together in an iterative fashion to strengthen the lesion harvesting at each round. Importantly, for (1) and (3) our framework can accept any state-of-the-art detector and classifier, allowing it to benefit from future improvements in these two domains. Even so, in this work, we develop our own \ac{LPG}, called \ac{3DCCN}, that combines the recent innovations seen in CenterNet~\cite{DBLP:journals/corr/abs-1904-07850} and the \ac{MULAN}~\cite{ieee/miccai/YanTPSBLS19}. We also propose a \ac{HNSL} to boost our \ac{LPG} with harvested hard negative cases. Among choices of \acp{LPC}, we use a \ac{MVCNN} to further reduce the false positive rate of produced lesion proposals. With our framework, we are able to harvest an additional $9,805$ lesions from DeepLesion, while keeping the label precision above $90\%$, at a cost of only fully annotating $5\%$ of the data. Compared to the original dataset this is a boost of $11.2\%$ in recall rates. Thus, our lesion harvesting framework, along with the introduced \ac{3DCCN} \ac{LPG} and \ac{HNSL}, represent our main contributions. 

However, we also provide additional important contributions. For one, we completely annotate and publicly release  $1\,915$ of the DeepLesion subvolumes with \ac{RECIST} marks, in addition to our harvested lesions. Second, we introduce and validate a new \ac{P3D} evaluation metric, designed for completely annotated data, that serves as a much better measurement of 3D detection performance than current practices\footnote{We open source our evaluation code, annotations, and results at \url{https://github.com/JimmyCai91/DeepLesionAnnotation.}}. Since all DeepLesion test data, up to this point, is incompletely annotated, these contributions allow for more accurate evaluations of lesion detection systems. Finally, we report concrete benefits of harvesting unlabeled lesions. To do this, we train several state-of-the-art detection systems~\cite{DBLP:journals/corr/abs-1904-07850,DBLP:journals/pami/HeGDG20} using data augmented with our harvested prospective lesions and hard negative examples. We show that with even the best published method to date~\cite{ieee/miccai/YanTPSBLS19}, the \ac{AP} can be improved by $\mathit{10}$ percent. We also show that our \ac{3DCCN} \ac{LPG} can also be used as an extremely effective detector, outperforming the state-of-the-art and providing additional methodological contributions to the lesion detection topic.

\section{Related Work} \label{sec:related}
\begin{figure*}[t!]
    \centering
    \includegraphics[width=\linewidth]{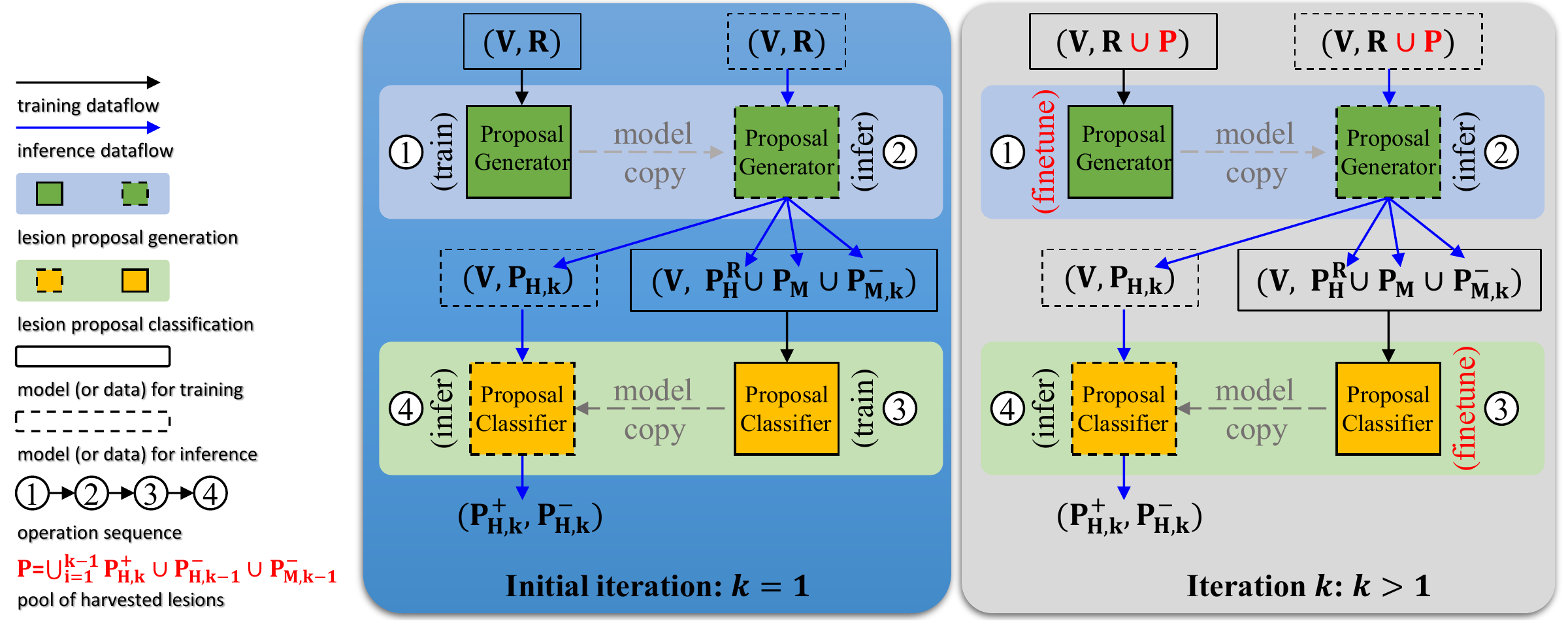}
    \caption{The flowchart of Lesion-Harvester. The second and last columns depict the initial iteration and follow-up iterations, respectively, where we use \textcircled{1}, \textcircled{2}, \textcircled{3}, and \textcircled{4} to indicate the sequence of operations. In step \textcircled{1}, we train (or finetune) the \acf{LPG}, and then, in step \textcircled{2}, we apply it on \ac{CT} volumes $\mathbf{V}$ to generate 3D proposals. In step \textcircled{3}, we train/finetune the \acf{LPC}, and apply it in step \textcircled{4} to automatically separate proposals into positive and negative groups as $\mathbf{P_{H,k}^+}$ and $\mathbf{P_{H,k}^-}$, respectively.}
    \label{fig:2}
\end{figure*}

\subsection{Detection with incomplete ground-truth.} 
The problem of missing annotations~\cite{DBLP:journals/pami/RenHG017,DBLP:conf/bmvc/WuBSNCD19} is related to, but differs from scenarios where the labeled data is independent from the unlabeled data. Thus, approaches to address the latter, such as deep growing learning~\cite{DBLP:conf/iccv/WangXLZ17}, which uses a small-sized but fully-annotated training set to initialize the model and then gradually expand training examples with pseudo labels, would not fully exploit DeepLesion labels. Instead, labeled and unlabeled lesions in DeepLesion often co-exist in the same training images, which changes the nature of the problem.

Ren \etal{},~\cite{DBLP:journals/pami/RenHG017} addressed the partial label problem and reduced the effect of missing labels by only using true positives and hard negatives for training detectors, where the latter were defined as region proposals with at least an overlap of $0.1$ with existing ground truth boxes. This setup, which we denote \acf{OBHS}, should help mitigate false negatives; however, it inevitably sacrifices a large amount of informative true negatives that have $\le$ $0.1$ overlap with ground truth boxes. Wu \etal{},~\cite{DBLP:conf/bmvc/WuBSNCD19} proposed \ac{OBSS} to improve object detectors, which weights contributions of region proposals proportional to their overlap with the ground truth. These strategies reduce the impact of true negatives not overlapping with the ground truth. Yet, true negatives in the background body structures are usually informative for training a robust lesion detector. By explicitly attempting to harvest prospective positives, Lesion-Harvester minimizes false negatives without suppressing informative background regions. 

Focusing specifically on DeepLesion, Wang \etal{},~\cite{DBLP:conf/prcv/WangLZZH19} first applied \ac{MGTM} to mine unlabeled lesions on the \ac{RECIST}-marked CT slices. Complementary with \ac{MGTM}, they also apply \ac{SLLP} to propagate bounding boxes from the \ac{RECIST}-marked slices to adjacent slices to reconstruct lesions in 3D. Different from \ac{MGTM}-\ac{SLLP}, Lesion-Harvester conducts lesion mining on the whole CT volume. It also alternately mines and then updates the deep learning models, to iteratively strengthen the harvesting process. Finally, it also identifies and then integrates hard negative examples using \acf{HNSL}.

\subsection{Label propagation from partial labels.}
Our work also relates to efforts on knowledge distillation and self label propagation. Radosavovic \etal{},~\cite{DBLP:conf/cvpr/RadosavovicDGGH18} proposed a data distillation method to ensemble predictions and automatically generate new annotations for unlabeled data from internet-scale data sources. Gao \etal{},~\cite{DBLP:conf/isbi/GaoXLWNSM16} investigated propagating labels from fully-supervised interstitial lung disease masks to unlabeled slices using \acp{CNN} and conditional random fields. Cai \etal{},~\cite{DBLP:conf/miccai/CaiTLHYXYS18} recovered 3D segmentation masks from 2D \ac{RECIST} marks in DeepLesion by integrating a \ac{CNN} and GrabCut~\cite{DBLP:journals/tog/RotherKB04}. We also tackle the large-scale and  noisy DeepLesion dataset, but we process each \ac{CT} volume as a whole, rather than focus on post-processing a given region of interest.

\subsection{Lesion detection.}
Recently, Yan \etal{},~\cite{yan_2018_deeplesion} introduced the DeepLesion dataset, which is a large-scale clinical database for whole body lesion detection, with follow-up work focusing on incorporating 3D context into 2D two-stage region-proposal \acp{CNN}~\cite{ieee/miccai/YanBS18,ieee/miccai/YanTPSBLS19}. These studies demonstrate the importance of incorporating 3D context in lesion detection and currently represent the state-of-the-art performance on DeepLesion. Some recent works also investigated one-stage detectors~\cite{ieee/miccai/ShaoGMLZ19,ieee/miccai/ZlochaDG19}. Compared with two-stage detectors, one-stage detectors are more flexible, straightforward, and computationally efficient. Recent work has also focused on false-positive detection. For instance, Ding \etal{},~\cite{ieee/miccai/DingLHW17} used a 3D-\ac{CNN} for this task, while Dou \etal{},~\cite{DBLP:journals/tbe/DouCYQH17} used a multi-scale context which ensembles three 3D-\acp{CNN} with small, medium, and large input patches. Varghese \etal{},~\cite{DBLP:journals/corr/VargheseVTKK16} proposed an interesting novelty detector (ND) for false positive reduction and applied it to brain tumor detection. More study is required to determine if their single layer denoising autoencoder has enough capacity to handle lesions with large appearance variabilities. Finally, Tang \etal{},~\cite{DBLP:conf/isbi/TangYTLXS19} showed that hard negative mining can increase detection sensitivity on DeepLesion. 

Unlike the above works, our main focus is on harvesting missing \textit{annotations} using a small amount of physician labor. To do so, we articulate the Lesion-Harvester pipeline that integrates state-of-the-art detection and classification solutions as \ac{LPG} and \ac{LPC}, respectively. We also introduce a one-stage multi-slice \ac{LPG} that performs better than prior work. Like Tang \etal{},~\cite{DBLP:conf/isbi/TangYTLXS19}, we also demonstrate the importance of hard negative mining. Finally, with our lesion harvesting task completed, we then show how the complete labels can be used to train the same \ac{LPG} detection frameworks to better localize lesions on \textit{unseen} data. 

\section{Methods} \label{sec:method}
\Fig~\ref{fig:2} overviews our proposed Lesion-Harvester. As motivated above, we aim to harvest missing annotations from the incomplete DeepLesion dataset~\cite{yan_2018_deeplesion,DBLP:conf/cvpr/YanWLZHBS18}. Additionally, as \Fig~\ref{fig:0}(c) demonstrates, another important aim is to fully localize the 3D extent of lesions, which means we aim to also generate 3D proposals for both \ac{RECIST}-marked and unlabeled lesions. In this section, we first overview our method in \Sec~\ref{sec:overview} and then detail each component in \Sec{}s~\ref{sec:generator}--\ref{sec:iterative-updating}. This is followed by \Sec~\ref{sec:p3d_method}, which outlines our proposed \acf{P3D} evaluation metric. 

\subsection{Overview} \label{sec:overview}

\textbf{For the problem setup}, we assume we are given a set of \ac{CT} volumes, $\mathbf{V}=\{v_i\}$. Within each volume, there is a set of \ac{RECIST}-marked lesions which are denoted as 
\begin{equation}
\mathbf{L^{R}_{i}}=\{\ell^{R}_{i,j}\}, 
\end{equation}
where $\ell^{R}_{i,j}$ is the $j$-th \ac{RECIST}-marked lesion in $v_i$. We would like to use 3D bounding boxes to represent each lesion or lesion proposal. However, because a \ac{RECIST} mark is only applied on the key slice, they only define a set of 2D boxes:
\begin{equation}
\mathbf{R}_{i}=\{r^{R}_{i,j}\}. 
\end{equation}
Because volumes are incompletely labeled, if we denote all lesions within volume $v_i$ as $\mathbf{L_i}$, then $\mathbf{L^R_i}$ is a subset of $\mathbf{L_i}$. Our goal is to both determine the 3D extent of unlabeled lesions:
\begin{equation}
\mathbf{L_i^U}=\mathbf{L_i}\setminus\mathbf{L_i^R},
\end{equation}
and also recover the full 3D extent of any \ac{RECIST}-marked lesion, \ie{}, convert $\mathbf{R}_{i}$ to 3D bounding boxes. To do this, we first construct a \textit{completely} annotated set of \ac{CT} volumes, $\mathbf{V_M}$, by augmenting the original \ac{RECIST} marks for these volumes with additional \textbf{manual} annotations for $\mathbf{L_M^U}$. We denote the supplementary and complete \ac{RECIST} marks as $\mathbf{R_M^U}$ and $\mathbf{R_M}$.

The remainder of volumes we wish to harvest from are denoted as $\mathbf{V_H}=\mathbf{V}\setminus\mathbf{V_M}$, which are accompanied by their \emph{incomplete} set of \ac{RECIST} marks, $\mathbf{R_H}$. By exploiting $\mathbf{R_M}$ and $\mathbf{R_H}$, we attempt to discover all unlabeled lesions in $\mathbf{V_H}$ and the full 3D extent of both marked and unmarked lesions. Importantly, we constrain the size of $\mathbf{V_M}$ to be much smaller than $\mathbf{V_H}$, \eg{}, $5\%$, to keep labor costs low.

\textbf{In the initial round,} we train a \acf{LPG} using only the \ac{RECIST}-derived 2D bounding boxes $\mathbf{R_M}$ and $\mathbf{R_H}$. To ensure flexibility, any state-of-the-art lesion detector can be used, either an off-the-shelf variant or the customized \acf{3DCCN} approach we elaborate in \Sec~\ref{sec:generator}. After convergence, we then execute the trained \ac{LPG} on $\mathbf{V}$, producing a set of 3D lesion proposals, $\mathbf{P}$. These likely cover a large number of lesions but they may suffer from high false positive rates. To correct this, we divide $\mathbf{P}$ into $\mathbf{P_M}$ and $\mathbf{P_H}$ as proposals generated from $\mathbf{V_M}$ and $\mathbf{V_H}$, respectively. By comparing $\mathbf{P_M}$ with $\mathbf{R_M}$, we can further divide it into true and false positives:
\begin{equation}
\mathbf{P_M^{R}}=\mathbf{P_M}\cap\mathbf{R_M}, \quad \textrm{and} \quad \mathbf{P_M^{-}}=\mathbf{P_M}\setminus\mathbf{R_M},
\end{equation}
where we use a pseudo-3D metric described in \Sec~\ref{sec:p3d_method} to determine whether a 3D proposal and a \ac{RECIST}-derived 2D bounding box intersect. Similarly, we can create a set of generated lesion proposals that intersect with the \ac{RECIST} marks from $\mathbf{V_H}$, which we denote $\mathbf{P_H^{R}}$. We then train a \acf{LPC} by using $\mathbf{P_H^{R}}$ and $\mathbf{P_M^{R}}$ for positive training examples and $\mathbf{P_M^{-}}$ for negative examples. Like the \ac{LPG}, any well-performing classification method can be used; however, we show that \acf{MVCNN} is particularly useful. The trained \ac{LPC} is then used to classify the remaining proposals from $\mathbf{P_H}$ into $\mathbf{P_H^{+}}$ and $\mathbf{P_H^{-}}$, which are the harvested positive and negative lesion proposals, respectively.

\textbf{In subsequent rounds,} we harvest positive and negative 3D proposals to finetune the \ac{LPG} and begin the process anew. Yet, when fine tuning the proposed \ac{3DCCN} we employ a \acf{HNSL} using $\mathbf{P_H^{-}}$ and $\mathbf{P_M^{-}}$ as mined hard negatives. We index algorithm iterations with subscript $k$ and each iteration will provide harvested lesion proposals. Accordingly, the complete pool of all harvested lesions is iteratively updated as
\begin{equation}
\mathbf{P_{H}^{+}}=\mathbf{P_{H}^{+}}\cup\mathbf{P_{H,k}^{+}},
\end{equation}  
\emph{where we abuse notation here} and use $\cup$ as an operator that will fuse lesion proposals of the same lesion by simply keeping the one with the highest detection confidence. For $\mathbf{P_{H}^{R}}$ we only keep proposals intersecting with the \ac{RECIST} marks. As for hard negatives, these are \emph{reset} after each iteration. Unless needed, we drop the round index $k$ for clarity. Below, we elaborate further on the individual system components.

\subsection{Lesion Proposal Generation} \label{sec:generator} 
\begin{figure}
    \centering 
    \includegraphics[width=\linewidth]{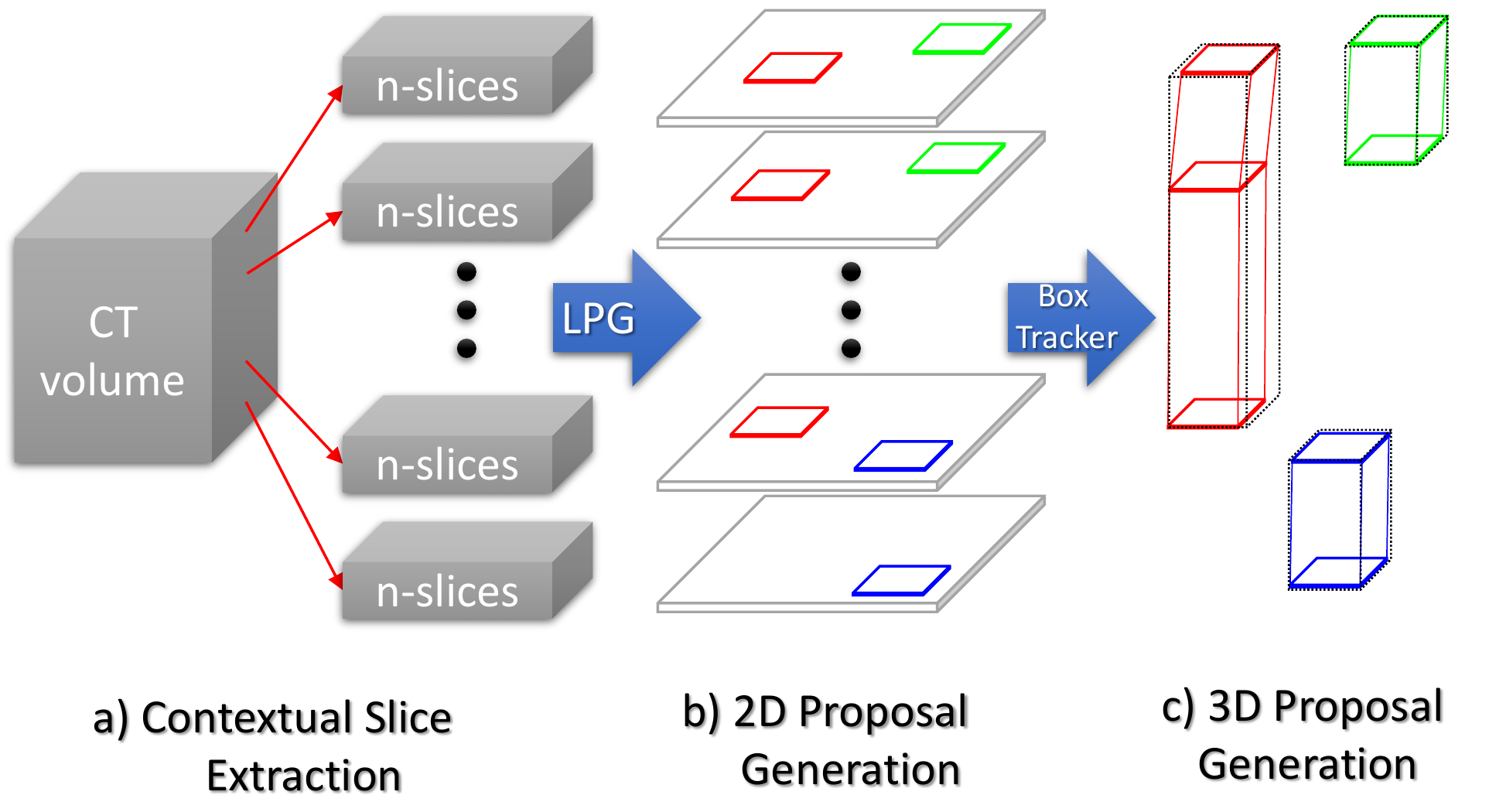} 
    \caption{Lesion proposal generation.} \label{fig:lpg}
\end{figure}

The \acf{LPG} uses a detection framework to generate lesion proposals. Following the state-of-the-art on DeepLesion~\cite{ieee/miccai/YanTPSBLS19}, our \ac{LPG} relies on a 2.5D lesion detection model to process CT images slice by slice; thus, 2D proposals must be aggregated together to produce 3D proposals. We outline each consideration below.

\subsubsection{\Acf{3DCCN}}
The task of our \ac{LPG} is to produce as high-quality lesion candidates as possible. While any state-of-the-art detection system can serve as \ac{LPG}, there are attributes which are beneficial. An \ac{LPG} with high sensitivity will help recover unlabeled lesions. Meanwhile, if it retains reasonable specificity, it will make downstream classification of proposals into true- and false-positives much more feasible. Computational efficiency is also important, to not only make training scalable, but also to be efficient in processing large amounts of \ac{CT} images. Finally, simplicity and efficiency are also crucial virtues, as the \ac{LPG} will be one component in a larger system. 

While 3D \acp{LPG} can be powerful, there is no straightforward approach to apply them on \ac{PACS}-mined data, like DeepLesion, which only provide 2D \ac{RECIST}-derived annotations. The \emph{de facto} standard in lesion detection for DeepLesion are 2D/2.5D approaches~\cite{yan_2018_deeplesion,ieee/miccai/YanBS18,ieee/miccai/YanTPSBLS19,DBLP:conf/isbi/TangYTLXS19,ieee/miccai/ZlochaDG19,ieee/miccai/ShaoGMLZ19}, which also avoid the prohibitive computational and memory demands of 3D \acp{LPG}. Thus, our \ac{LPG} of choice is a \ac{3DCCN}, which combines state-of-the-art one-stage anchor-free 2D proposal generation~\cite{DBLP:journals/corr/abs-1904-07850} with 3D context fusion~\cite{ieee/miccai/YanTPSBLS19}. Others have articulated the benefits of dense pixel-wise supervision~\cite{DBLP:conf/isbi/TangYTLXS19,DBLP:journals/pami/HeGDG20} and one-stage approaches provide a more straightforward means to aggregate such signals. Additionally, the choice of an anchor-free approach avoids the need to tune anchor-related hyper-parameters. Because lesions have convex shapes which have centroids located inside lesions, the center-based loss of CenterNet~\cite{DBLP:journals/corr/abs-1904-07850} is a natural choice. A final advantage to the one-stage anchor-free approach, which we will show in \Sec~\ref{sec:iterative-updating}, is that it allows for a natural incorporation of hard negative examples, which significantly improves performance. 

We then follow the same pipeline and hyper-parameter settings described by Zhou \etal{},~\cite{DBLP:journals/corr/abs-1904-07850}. Namely, we create ground-truth heat-maps centered at each lesion using Gaussian kernels ${Y \in [0,1]^{\tilde{W} \times \tilde{H}}}$. The training objective is to then produce a heatmap, $\hat{Y}_{xy}$, using a penalty-reduced pixel-wise logistic regression with focal loss~\cite{DBLP:conf/iccv/LinGGHD17}:
\begin{equation} \label{eq:center-loss}
    E_{\ac{LPG}} = \frac{-1}{m} \sum_{xy}
    \begin{cases}
        (1 - \hat{Y}_{xy})^{\alpha} 
        \log(\hat{Y}_{xy}) & \!\text{if}\ Y_{xy}=1\vspace{2mm}\\
        \begin{array}{c}
        (1-Y_{xy})^{\beta} 
        (\hat{Y}_{xy})^{\alpha}\\
        \log(1-\hat{Y}_{xy})
        \end{array}
        & \!\text{otherwise}
    \end{cases} \textrm{,}
\end{equation}
where $m$ is the number of objects in the slice and $\alpha=2$ and $\beta=4$ are hyper-parameters of the center loss. At every output pixel, the width, height, and offset of lesions are also regressed, but they are only supervised where $Y_{xy}=1$. The lesion proposals are produced by combining center points with regressed width and height. See Zhou \etal{},~\cite{DBLP:journals/corr/abs-1904-07850}~for more details. 

To incorporate 3D context, which Yan \etal{},~\cite{ieee/miccai/YanTPSBLS19} demonstrated can benefit lesion detection, we use a 2.5D DenseNet-121 backbone~\cite{DBLP:conf/cvpr/HuangLMW17}. This backbone is associated with the highest performance for DeepLesion detection to-date~\cite{ieee/miccai/YanTPSBLS19} and functions by including consecutive \ac{CT} slices as input channels. Based on a balance between performance and computational efficiency, we follow Yan \etal{},~\cite{ieee/miccai/YanTPSBLS19} choose to include the $4$ adjoining slices above and below.

\subsubsection{3D Proposal Generation}
Regardless of the \ac{LPG} used, if it operates slice-wise, like the \ac{3DCCN}, then post-processing is required to generate 3D proposals, $\mathbf{P}$. To do this, we first apply the \ac{LPG} to scan over \ac{CT} volumes generating detection results on every axial slice. This produces a set of 2D proposals, each with a detection score. Next, we stack proposals in consecutive slices using the same Kalman filter-based bounding box tracker as Yang \etal{},~\cite{DBLP:journals/access/YangCLLWY19}. More specifically, we first select 2D proposals whose detection score is greater than a threshold $t_{G}$. 2D proposals from adjoining slices are then stacked together if their \ac{IoU} is $\ge 0.8$. Finally, in case the \ac{LPG} misses lesions in intermediate slices, we extend each 3D box up and down by one slice, and if any two 3D boxes become connected with $\ge 0.8$ overlap on the connecting slices, then they will be fused as one 3D proposal. When lesion harvesting, we choose $0.1$ as the value for $t_{G}$ which helps keep the number of proposals manageable. However, our experience indicates that results are not sensitive to significant deviations from our chosen threshold value. 

The next step in our process is to separate lesion candidates, $\mathbf{P}$, into true- and false-positives. Because we have access to a small subset of fully-annotated volumes, $\mathbf{V_M}$, we can identify the 3D proposals that overlap (see \Sec~\ref{sec:p3d_method}) with the \ac{RECIST} marks to be true-positives and denote them as $\mathbf{P_M^R}$. The remaining false-positive proposals are denoted $\mathbf{P_M^{-}}$. We can also identify true-positives $\mathbf{P_H^R}$ that overlap with existing \ac{RECIST} marks in $\mathbf{V_H}$. Because $\mathbf{V_H}$ is only partially labeled, the remainder of proposals must be filtered somehow into true and false positives. 

\subsection{Lesion Proposal Classification}  

\begin{figure}
    \centering
    \includegraphics[width=\linewidth]{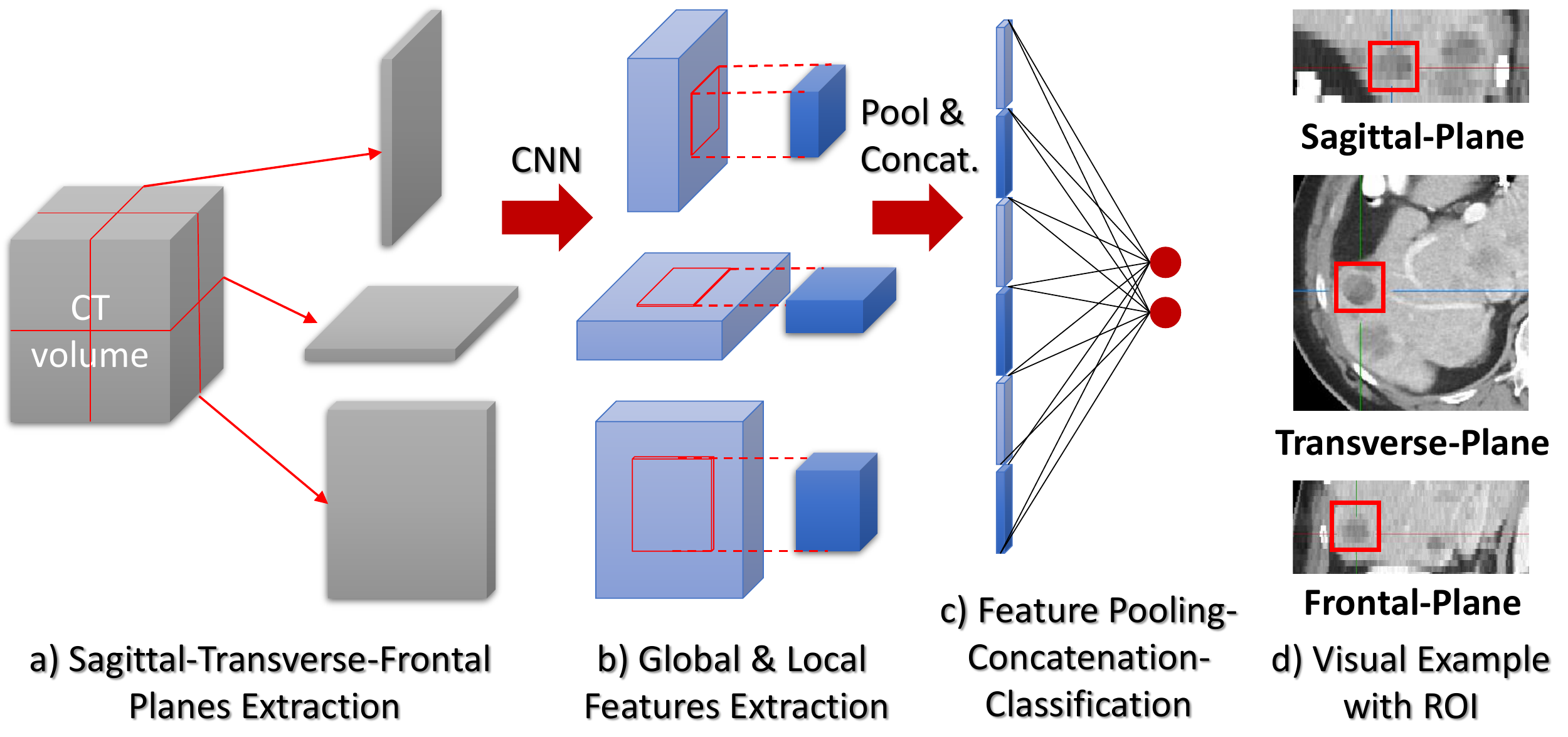} 
    \caption{Lesion proposal classification using \acf{MVCNN}.} \label{fig:lpc}
\end{figure}

With the manually verified proposals in hand, namely $\mathbf{P_M^R}$, $\mathbf{P_M^{-}}$, and $\mathbf{P_H^R}$, the aim is to identify proposals in $\mathbf{P_H}\setminus\mathbf{P_H^R}$. To do this, we use the verified proposals to train a binary \acf{LPC}. In principle, any classifier can be used, but we opt for a \acf{MVCNN}, which combines two main concepts. The first concept is that 3D context is necessary for differentiating true positive lesion proposals from false positives~\cite{DBLP:journals/tbe/DouCYQH17,ieee/miccai/DingLHW17}, whether for machines or for clinicians. A fully 3D classifier can satisfy this need, but, as we show in the results, a multi-view approach that operates on transverse, sagittal, and coronal planes centered at each proposal can perform better. This matches prior practices~\cite{ieee/miccai/RothLSCHWLTS14} and has the virtue of offering a much more computational and memory efficient means to encode 3D context compared to true 3D networks.

The second concept is multi-scale learning. As reported by Yan \etal{},~\cite{DBLP:conf/cvpr/YanWLZHBS18}, a  ``global'' context can aid in lesion characterization. This is based on the intuition that the surrounding anatomy can help place a prior on how lesions should appear. We use the same recommendation as Yan \etal{}, and use a $128\times128\times32$ region centered at each lesion to extract global features. ROI pooling based on the lesion proposal is then used to extract more local features. The local and global features are concatenated together before being processed by a fully-connected layer. The \ac{MVCNN} is shown in \Fig~\ref{fig:lpc}.

We choose to use a ResNet-18~\cite{DBLP:conf/cvpr/HeZRS16} as our backbone because of its proven usefulness and availability of pre-trained weights. The \ac{LPC} is trained with the manually verified proposals using cross-entropy loss. We expect $\mathbf{P_M^R}$, $\mathbf{P_M^{-}}$, and $\mathbf{P_H^R}$ to be representative of the actual distribution of lesions in DeepLesion. In particular, although the negative samples $\mathbf{P_M^{-}}$ are only generated from $\mathbf{V_M}$, they should also be representative to the dataset-wide distribution of hard negatives since \emph{the hard negatives are typically healthy body structures, which are common across patients}. 

With the \ac{LPC} trained, we then apply it to the proposals needing harvesting: $\mathbf{P_H}\setminus\mathbf{P_H^R}$. Since the \ac{LPG} and \ac{LPC} are independently trained, we make an assumption, for simplicity, that their pseudo-probability outputs are independent as well. Thus, the final score of a 3D proposal can be calculated as 
\begin{equation}
    s_{\{p|g,c\}} = s_g \cdot s_c, \label{eqn:proposal_score}
\end{equation}
where $s_{\{p|g,c\}}$ is called the lesion score and $s_g$ and $s_c$ are the \ac{LPG} detection score and \ac{LPC} classification probability, respectively. We obtain the former by taking the max detection score across all 2D boxes in the proposal. Based on $s_{\{p|g,c\}}$, we generate prospective positive proposals, $\mathbf{P_H^{+}}$, by choosing a threshold for $s_{\{p|g,c\}}$ that corresponds to a precision above $95\%$ on the completely annotated set $\mathbf{V_M}$. \textcolor{black}{From the remainder, we select proposals whose detection score satisfy $s_g \ge 0.5$ as the hard negative examples. Then from each volume we choose up to five negative examples with the top detection scores to construct $\mathbf{P_H^{-}}$.}

\subsection{Iterative Updating} \label{sec:iterative-updating}
After a round of harvesting, we repeat the process by fine-tuning the \ac{LPG}, but with important differences. First, we now have prospective positive proposals corresponding to unlabeled lesions, \ie{}, $\mathbf{P_H^{+}}$, to feed into training. In addition, for all proposals, even those corresponding to \ac{RECIST}-marked lesions, we now have 3D proposals. To keep computational demands reasonable, only the 2D slices with the highest detection score within each proposal are used as additional samples to augment the original \ac{RECIST} slices.

\begin{figure}[t!]
    \centering
    \includegraphics[width=.98\linewidth]{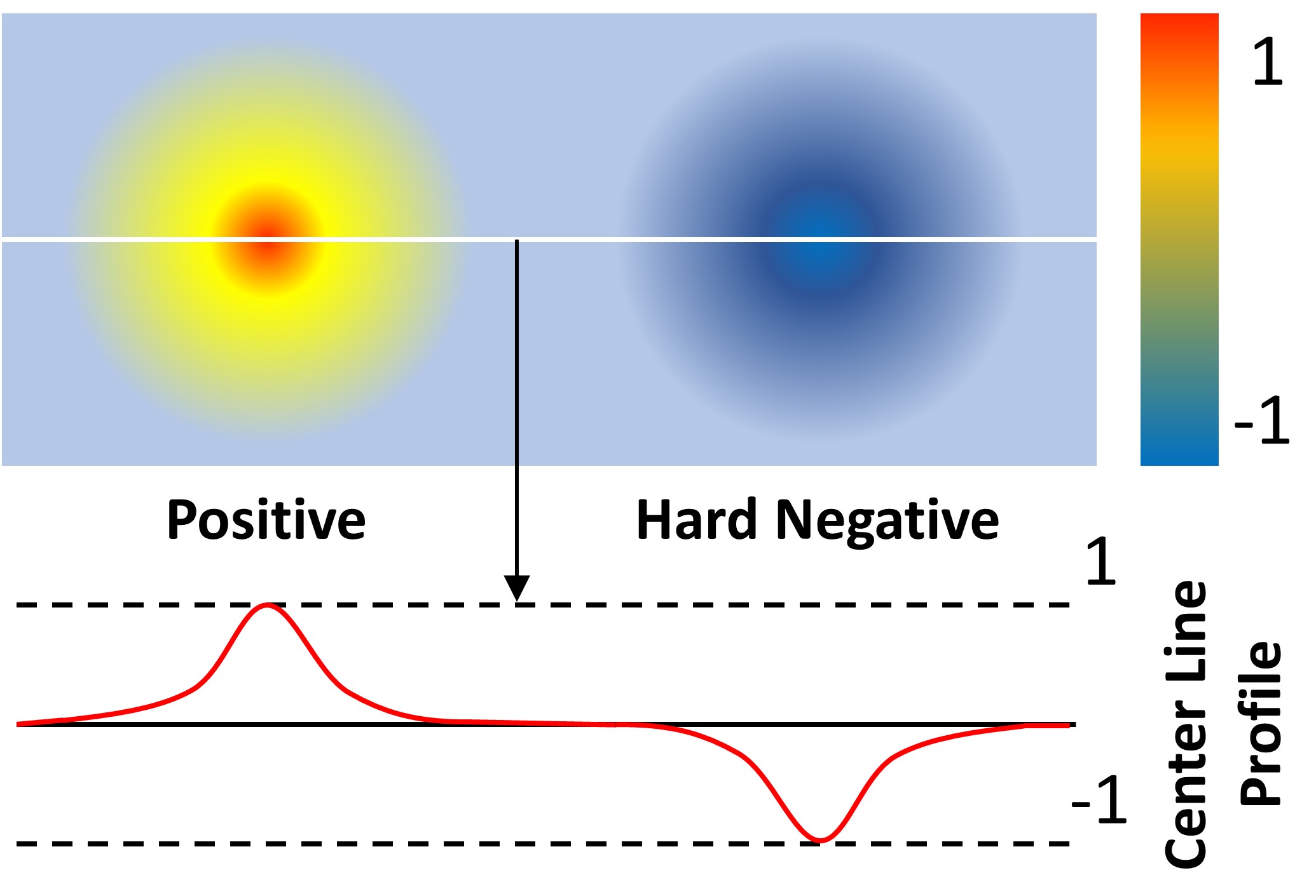}  
    \caption{Ground-truth heatmaps for positives and hard-negative examples are shown in the left and right, respectively. We define both \ac{RECIST}-marked lesions and mined lesions to be positive examples.}
    \label{fig:heatmaps}
\end{figure}

Secondly, to incorporate harvested hard negative proposals, we use the same procedure in \Sec~\ref{sec:generator}, but replace the center-loss in \Eq~\ref{eq:center-loss} with our proposed \acf{HNSL}. To do this, we create separate heat maps for positive (\ac{RECIST}-marked or prospective positive) and hard-negative lesions. We denote these heat maps as $Y_{xy}^{p}$ and $Y_{xy}^{n}$, respectively. We then create a master ground truth heat map, $Y_{xy}$, by overwriting $Y_{xy}^{p}$ with $Y_{xy}^{n}$:
\begin{equation} \label{eq:center-loss2}
    Y_{xy} = 
    \begin{cases}
        -Y_{xy}^{n} & \!\text{if}\ Y_{xy}^{n}>0\vspace{2mm}\\
        Y_{xy}^{p}
        & \!\text{otherwise}
    \end{cases} \textrm{.}
\end{equation}
The result is a ground truth map that can now range from $[-1, 1]$. When used in the loss of \eqref{eq:center-loss}, the effect is that positive predictions in hard negative regions are penalized much heavier than standard negative regions ($16$ times heavier when $\beta=4$). This simple  modification works surprisingly well for further reducing false positive rates. We visually depict example ground truth heatmaps in \Fig~\ref{fig:heatmaps}. 

\subsection{Pseudo-3D Evaluation}
\label{sec:p3d_method}

Apart from the lesion completion framework, introduced above, another important aspect to discuss is evaluation. Current DeepLesion works~\cite{yan_2018_deeplesion,ieee/miccai/YanBS18,ieee/miccai/YanTPSBLS19,DBLP:conf/isbi/TangYTLXS19,ieee/miccai/ZlochaDG19,ieee/miccai/ZlochaDG19,ieee/miccai/ShaoGMLZ19} operate and evaluate only based on the 2D \ac{RECIST} marks on selected 2D slices that happen to contain said marks. This is problematic, as \ac{RECIST}-based evaluation will not reflect actual performance: it will miscount \textit{true positives} on unmarked lesions or on adjoining slices as \textit{false positives}. Moreover, automated methods should process the whole image volume, meaning precision should be correlated to false positives per \textit{volume} rather than per selected \textit{slice}. In this way, automated methods can be more effective on holistically describing and recording tumor existence, complimentary to human efforts to better achieve precision medicine.

Because we aim to harvest 3D bounding boxes that cover all lesions, we must evaluate, by definition, on completely annotated test data. Yet, it is not realistic to assume data will be fully annotated with 3D bounding boxes. Instead, a more realistic prospect is that test data will be completely annotated with 2D \ac{RECIST} marks, especially by clinicians who are more accustomed to this. Thus, assuming this is the test data available, we propose a \acf{P3D} \ac{IoU} metric. For each \ac{RECIST} mark, we can generate 2D bounding boxes based off of their extent, as in~\cite{yan_2018_deeplesion}. This we denote  $\left(x_1,x_2,y_1,y_2,z,z\right)$, where $z$ is the slice containing the mark. Given a 3D bounding box proposal,  $\left(x'_1,x'_2,y'_1,y'_2,z'_1,z'_2\right)$, our \ac{P3D} IoU metric will be counted as a true positive if and only if $z'_1 \leq z \leq z'_2$ and $IoU\left[\left(x_1, x_2, y_1, y_2\right), \left(x'_1, x'_2, y'_1, y'_2\right)\right] \ge 0.5$. Otherwise, it is considered a false positive. Because we publicly release complete \ac{RECIST} marks of $1915$ volumes, the \ac{P3D} \ac{IoU} metric can also be used to benchmark DeepLesion detection performance, replacing the one currently used. As we show in the results, the \ac{P3D} \ac{IoU} metric is a much more accurate performance measure. 

\section{Experiments} \label{sec:experiments}

\subsection{Dataset}
\begin{figure}[t!]
\centering  
\includegraphics[width=.88\linewidth]{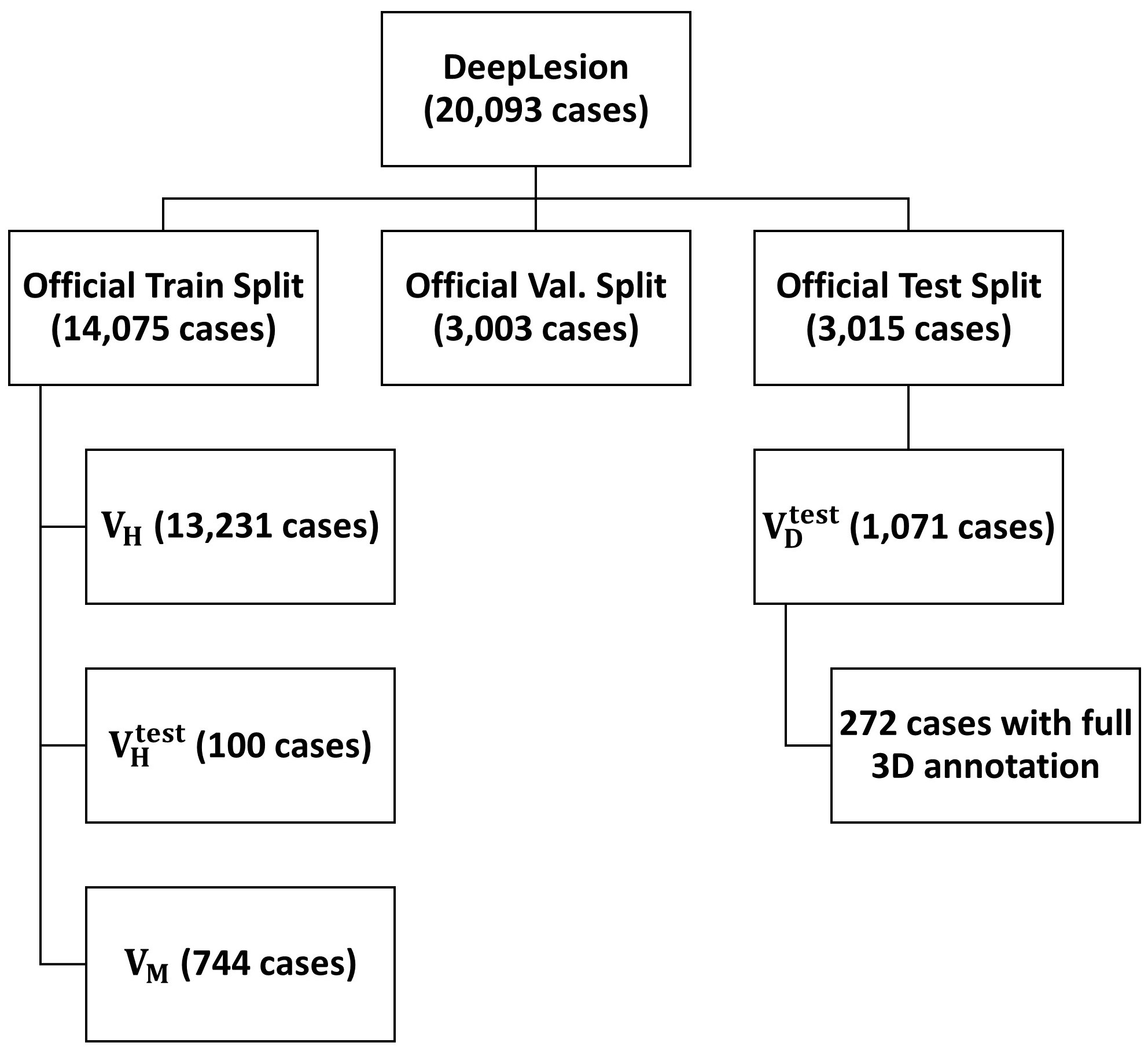}  
\caption{\textbf{Dataset Splits:} we follow DeepLesion's official data split and further define $\mathbf{V_H}$, $\mathbf{V_H^{test}}$, $\mathbf{V_M}$, and $\mathbf{V_D^{test}}$ for the needs of lesion harvesting.}
\label{fig:dataset-splits}
\end{figure}

To harvest lesions from the DeepLesion dataset, we randomly selected $844$ volumes from the original $14\,075$ training \acp{CT}\footnote{\url{https://nihcc.app.box.com/v/DeepLesion}}. These are then annotated by a board-certified radiologist. Of these, we select $744$ as $\mathbf{V_M}$ ($5.3\%$) and leave another $100$ as an evaluation set for lesion harvesting. This latter subset, denoted $\mathbf{V_H^{test}}$, is treated identically at $\mathbf{V_H}$, meaning the algorithm only sees the original DeepLesion \ac{RECIST} marks. After convergence, we can measure the precision and recall of the harvested lesions. In addition, we later measure detection performance on systems trained on our harvested lesions by also fully annotating, with \ac{RECIST} marks, $1,071$ of the testing \ac{CT} volumes. These volumes, denoted $\mathbf{V_D^{test}}$, are never seen in our harvesting framework. To clarify the dataset split in our experiments, we show it hierarchically in \Fig~\ref{fig:dataset-splits}.

\subsection{\ac{P3D} IoU Evaluation Metric}

\begin{figure}[t!]
    \centering 
    \subfloat[$0.838$, $p \ll 10^{-5}$.\label{exp:scatter-3d-recist-recall}]{
        \includegraphics[trim=0.2in 0.7in 1.4in 0.1in,clip,width=.40\linewidth,height=1.2in]{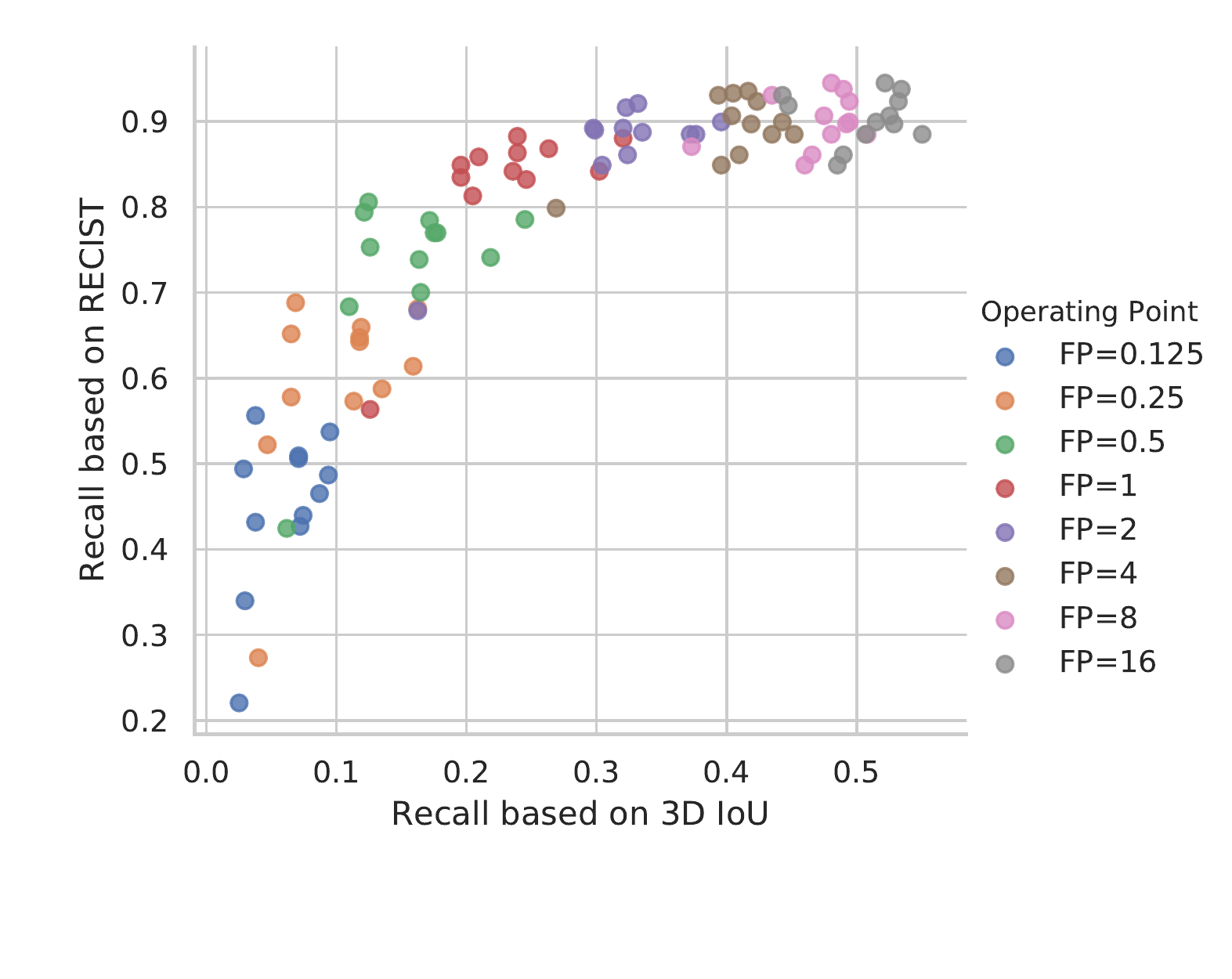}}
    \subfloat[$0.993$, $p \ll 10^{-5}$; \label{exp:scatter-3d-p3d-recall}]{
        \includegraphics[trim=0.2in 0.7in 0.1in 0.1in,clip,width=.55\linewidth,height=1.2in]{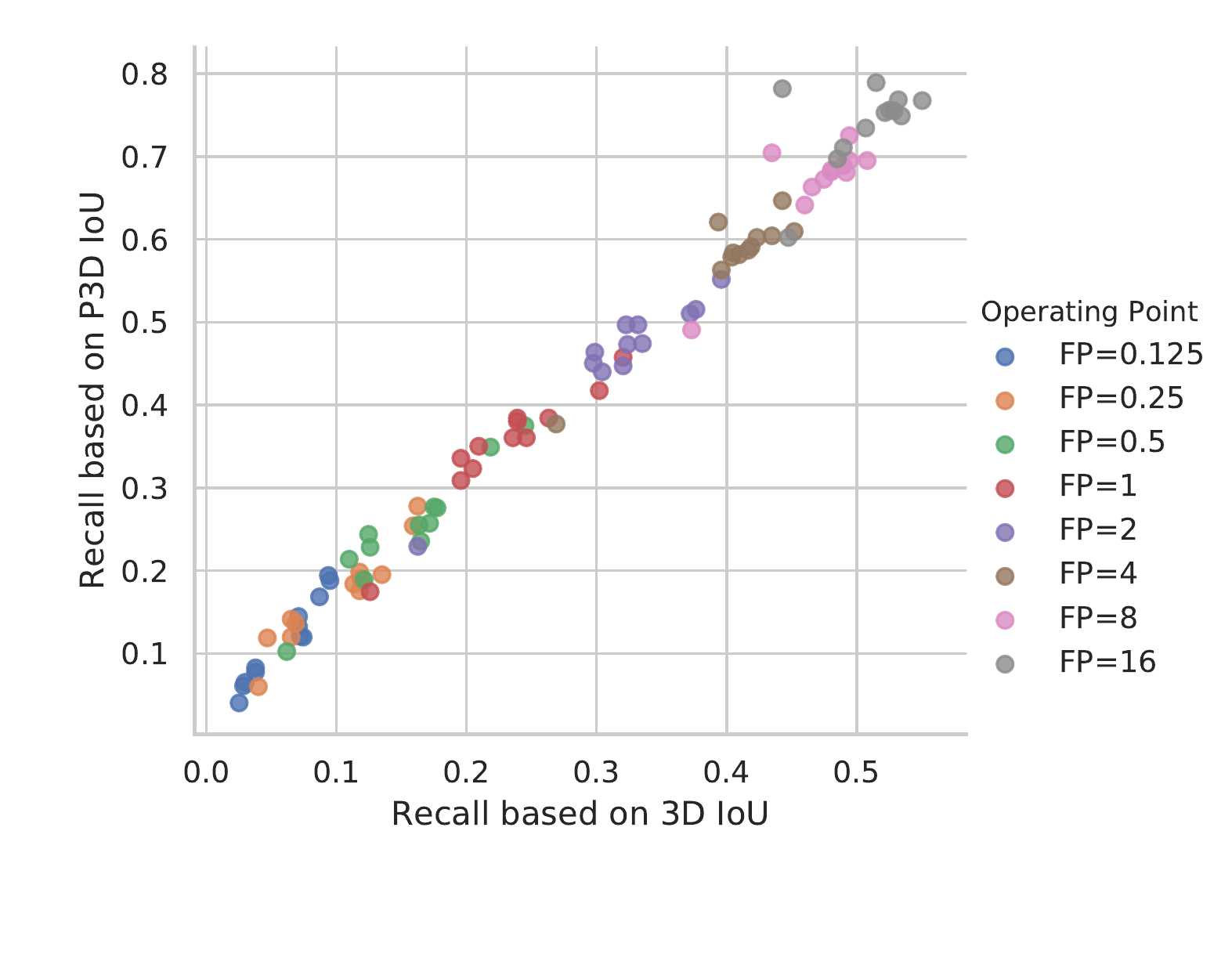}}
    \caption{Comparing the concordance of the incomplete 2D \ac{RECIST} and our \ac{P3D} metric compared to the gold-standard 3D IoU metric. For each metric, lesion detection recalls at operation points from FP=$0.125$ to FP=$16$ are collected from the \ac{FROC} curves of 12 lesion detection methods. Pearson coefficients are shown below each chart.}
\end{figure}

Before validating our lesion harvesting framework, we first validate our proposed \ac{P3D} metric. To do this, we used \textit{3D bounding boxes} to annotate a small set of $272$ CT test volumes, randomly selected from $\mathbf{V_D^{test}}$. From these, we can calculate a ``gold-standard'' 3D \ac{IoU} metric, and analyze the concordance of different proxies. Accordingly, we trained state-of-the art detection methods (the same $12$ outlined in \Tab~\ref{tab:detection3d}'s later experiments) on the DeepLesion dataset and measured their performance using 3D \ac{IoU}, incomplete \ac{RECIST}~\cite{yan_2018_deeplesion}, and the proposed \ac{P3D} \ac{IoU} metrics. \textcolor{black}{We use a 3D \ac{IoU} threshold} of $0.3$, instead of $0.5$ commonly used in 2D applications, to help compensate for the severity of 3D \ac{IoU}. As shown in \Fig~\ref{exp:scatter-3d-p3d-recall} and \Fig~\ref{exp:scatter-3d-recist-recall}, we measured \ac{FROC} curves and compare detection recalls of these methods at operating points varying from \ac{FP} rates of $0.125$ to $16$ per volume. As can be seen, our \ac{P3D} metric has much higher concordance with the true 3D \ac{IoU} than does the incomplete 2D \ac{RECIST} metric. Moreover, the latter exhibits a relationship that is much noisier and non-monotonic, making it likely any ranking of methods does not correspond to their true ranking. Thus, for the remainder of this work we report lesion harvesting and lesion detection performance using only the \ac{P3D} metric. Moreover, we advocate using the \ac{P3D} metric, and the fully \ac{RECIST}-annotated test sets we publicly release, to evaluate DeepLesion detection systems going forward. 

We also evaluated whether the Kalman filtering method in \Sec~\ref{sec:generator} can produce accurate 3D lesion proposals from 2D detections. Regardless of the detection framework used, median 3D \acp{IoU} are $0.4$, which is a high overlap for 3D detection. All of the first quartiles are above $0.2$ 3D \ac{IoU}, indicating that the most of the reconstructed 3D boxes are of high quality. Violin plots can be found in our supplementary material.

\subsection{Main Result: Lesion Harvesting}

\begin{figure*}[t!]
    \centering 
    \subfloat[\label{exp:fig-lesion-mining}]{
        \includegraphics[trim=0.05in 0.85in 0.4in 0.55in,clip,width=.48\linewidth, height=2.2in]{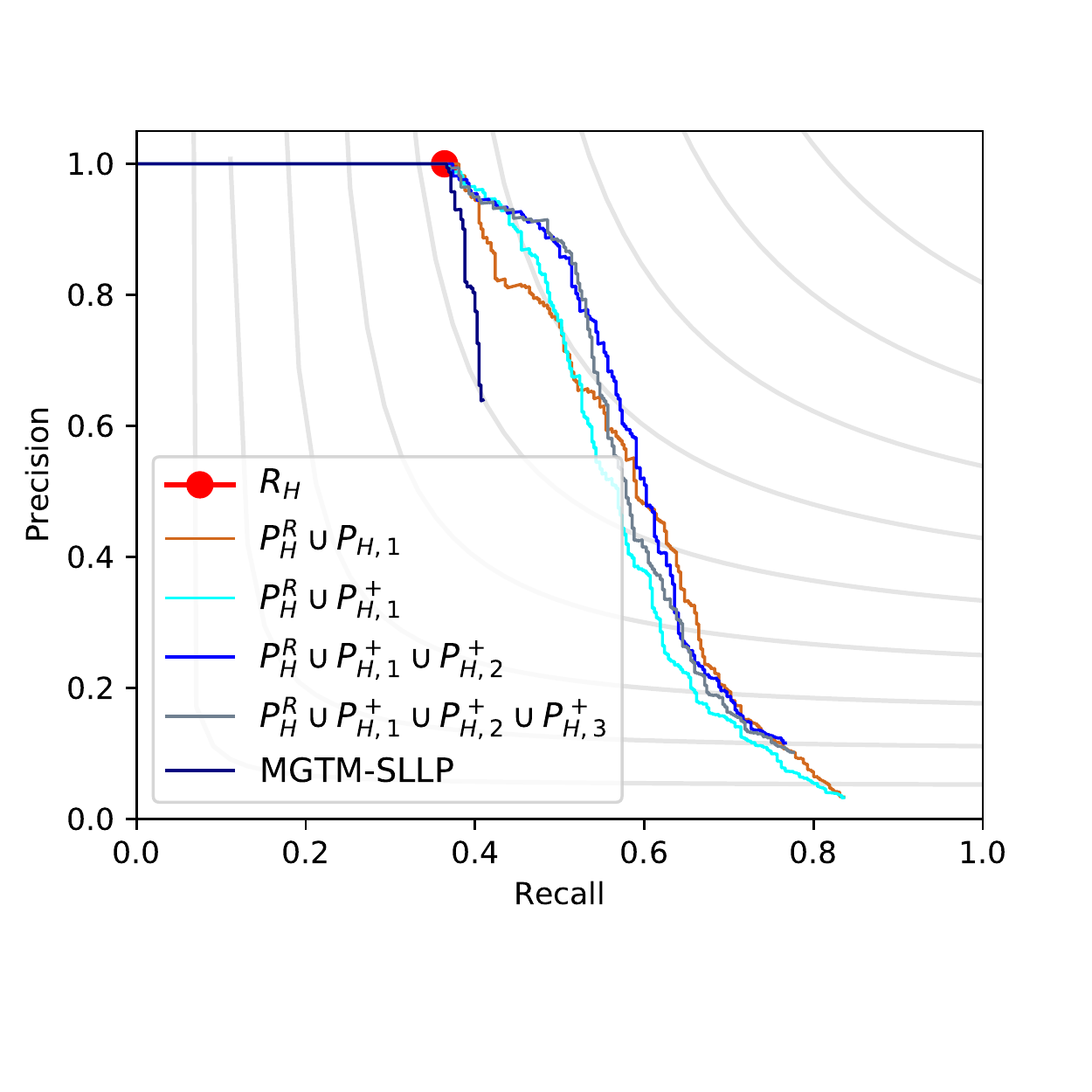}}
    \subfloat[\label{fig:exp-lpc}]{
        \includegraphics[trim=0in 0.8in 0.2in 0.54in,clip,width=.48\linewidth, height=2.2in]{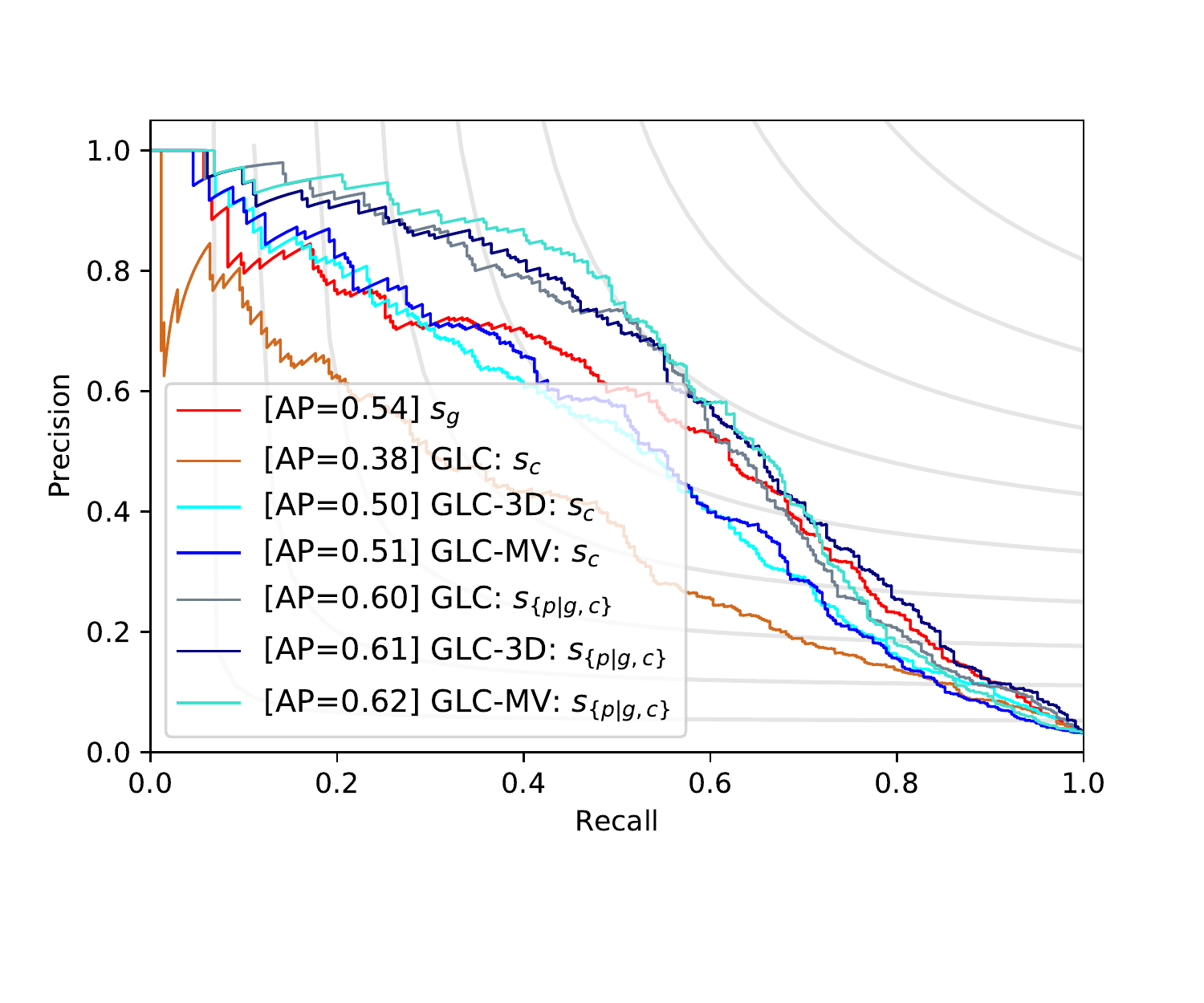}}
    \caption{(a) \ac{PR} curve evaluating our iterative lesion harvesting procedure. The original \ac{RECIST} marks only have a recall of $36.7\%$, which is shown as the red dot. $\mathbf{P_{H,1}}$ is the set of lesion proposals from \ac{LPG}, which are then filtered by the \ac{LPC} to generate $\mathbf{P_{H,1}^+}$. $\mathbf{P_{H,2}^+}$ and $\mathbf{P_{H,3}^+}$ are harvested lesions from the second and third iterations, respectively. \ac{MGTM}-\ac{SLLP}~\cite{DBLP:conf/prcv/WangLZZH19} is a baseline method for mining missing lesions. (b) \ac{PR} curves of different \acp{LPC} in the first iteration, where \ac{GLC}, \ac{GL3D}, and \ac{MVCNN} denote classifiers using the transverse-plane, the multi-view sagittal-transverse-frontal plane, and 3D sub-volume inputs, respectively.}
\end{figure*}

\begin{table}[t!]
    \centering
    \caption{Lesion harvesting performance evaluated on $\mathbf{V_H^{test}}$. Detection recalls (R) at precisions (P) from 80\% to 95\% are reported after six harvesting rounds.} 
    \label{exp:tab-lesion-mining}
    \begin{tabular}{lccccc}
        \toprule
        {\bf Training Label Set} & \acs{R@80P} & \acs{R@85P} & \acs{R@90P} & \acs{R@95P} & Avg. \\ 
        \midrule
        $\mathbf{R_H}$ (\ac{RECIST})                 & \multicolumn{5}{c}{36.7$@$100P} \\
        \acs{MGTM}-\acs{SLLP}~\cite{DBLP:conf/prcv/WangLZZH19} & 40.0 & 38.8 & 38.6 & 37.6 & 38.7 \\
        \midrule
        $\mathbf{P_H^R} \cup \mathbf{P_{H,1}}$                   & 46.7 & 42.4 & 40.7 & 39.9 & 42.4 \\
        $\mathbf{P_H^R} \cup \mathbf{P_{H,1}^+}$                  & 46.3 & 45.7 & 44.0 & 41.7 & 44.5 \\
        $\mathbf{P_H^R} \cup \bigcup_{i=1}^{2}\mathbf{P_{H,i}^+}$ & 52.4 & 50.0 & 44.7 & 41.3 & 47.1 \\
        $\mathbf{P_H^R} \cup \bigcup_{i=1}^{3}\mathbf{P_{H,i}^+}$ & 51.4 & \textbf{50.5} & \textbf{47.9} & 41.2 & 47.7 \\
        $\mathbf{P_H^R} \cup \bigcup_{i=1}^{4}\mathbf{P_{H,i}^+}$ & \textbf{53.3} & 50.0 & 46.1 & 41.9 & \textbf{47.8} \\
        $\mathbf{P_H^R} \cup \bigcup_{i=1}^{5}\mathbf{P_{H,i}^+}$ & 52.9 & 48.6 & 45.8 & \textbf{42.5} & 47.4 \\
        $\mathbf{P_H^R} \cup \bigcup_{i=1}^{6}\mathbf{P_{H,i}^+}$ & \textbf{53.3} & 48.6 & 43.8 & 40.2 & 46.5 \\
        \bottomrule
    \end{tabular}
\end{table}

We validate our lesion harvesting by running it for 6 iterations. As can be seen in \Tab~\ref{exp:tab-lesion-mining}, the set of original \ac{RECIST}-marked lesions, $\mathbf{R_H}$, only has a recall of $36.7\%$ for the lesions in $\mathbf{V_H^{test}}$, with an assumed precision of $100\%$. After one iteration, the initial lesion proposals generated by the \ac{3DCCN}-based \ac{LPG}, denoted as $\mathbf{P_{H,1}}$, can boost the recall to $40.7\%$, while keeping the precision at $90\%$. However, after filtering with our \ac{MVCNN}-based \ac{LPC}, which selects $\mathbf{P_{H,1}^{+}}$ from $\mathbf{P_{H,1}}$, the recall is boosted to $44.0\%$, representing $8\%$ increase in recall over $\mathbf{R_H}$. This demonstrates the power and usefulness of our \ac{LPG} and \ac{LPC} duo. After $3$ rounds of our system, the performance increases further, topping out at $47.9\%$ recall at $90\%$ precision. This corresponds to harvesting $\mathbf{9\,805}$ more lesions from the $\mathbf{V_H}$ CT volumes. As \Tab~\ref{exp:tab-lesion-mining} also indicates, running Lesion-Harvester for more rounds after $3$ does not notably contribute to performance improvements. \Fig~\ref{exp:fig-lesion-mining}, depicts \ac{PR} curves for the first three rounds and it can be seen that \ac{PR} curve has begun to top out after the third round. Finally, we also compare against the \ac{MGTM}-\ac{SLLP} lesion mining algorithm~\cite{DBLP:conf/prcv/WangLZZH19}. As can be seen, while \ac{MGTM}-\ac{SLLP} can also improve recall, the Lesion-Harvester outperforms it by $9.3\%$ in recall at $90\%$ precision.

\begin{figure*}[t!]
    \centering
    \includegraphics[trim=0in 2in 0in 0.2in,clip,width=.80\linewidth]{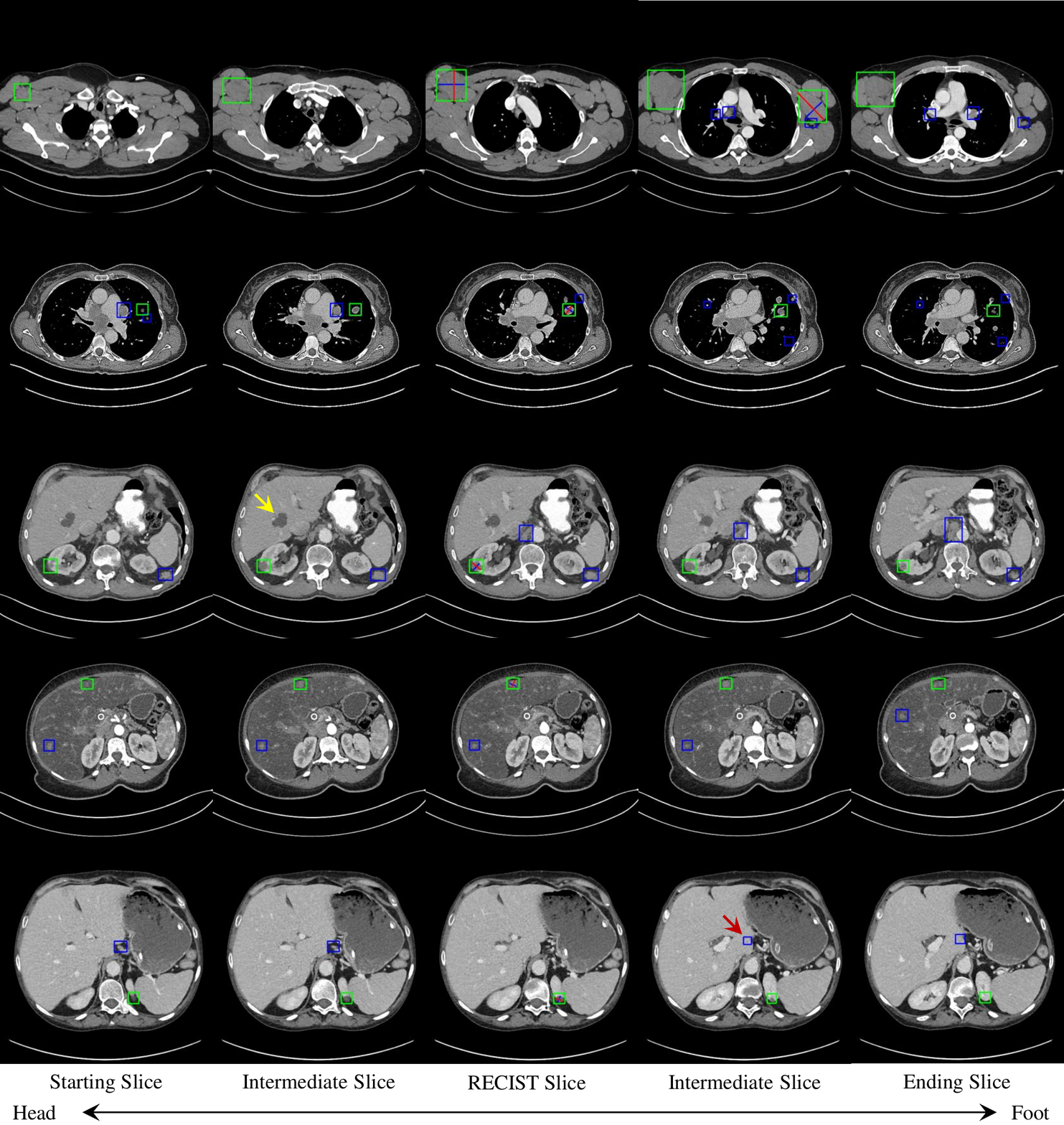}
    \caption{Examples of 3D detection results and mined positive lesions from the harvesting set $\mathbf{V_H}$. We use green and blue boxes to show \ac{RECIST}-marked and mined lesions, respectively. Each 3D detection consists of multiple axial slices and we show 5 typical slices: the starting slice, the \ac{RECIST} slice, the ending slice, and two intermediate slices. We show RECIST marks as crosses with red and blue lines. We also show one failure case at the bottom row indicated by the red arrow and a lesion still remains unlabeled at the 3rd row indicated by the yellow arrow.} 
    \label{fig:exp-vis}
\end{figure*}

Importantly all 2D lesion bounding boxes are now also converted to 3D. It should be stressed that these results are obtained by annotating $744$ volumes, which represents only $5.3\%$ of the original training split of DeepLesion. In terms of the distribution of harvested lesions, they match the original DeepLesion distribution which is weighted toward lung, liver, kidney lesions as well as enlarged lymph nodes. In our supplementary material, we illustrate the similarity of the distributions between the original DeepLesion lesions and the harvested ones in details. \Fig~\ref{fig:exp-vis} provides some visual examples of the harvested lesions. As can be seen, lesions missing from the original \ac{RECIST} marks can be harvested. These examples, coupled with the quantitative boosts in recall (seen in \Fig~\ref{exp:fig-lesion-mining}), demonstrate the utility and power of our lesion harvesting approach.

In our implementation, we trained the \ac{LPG} and \ac{LPC} using $3$ NVIDIA RTX6000 GPUs, which took a few hours to converge for each round. However, producing the lesion proposals after each round is time consuming ($\sim 12$ hours) since it requires the \ac{LPG} to scan every CT slice in DeepLesion. In total, the Lesion-Harvester takes $3$ days to converge on DeepLesion.

\subsection{Ablation Study: \ac{LPG}}

\begin{table*}[t!]
    \begin{center}
        \caption{\textcolor{black}{Evaluations of \ac{LPG} trained with different label sets including the original \ac{RECIST} marks, $\mathbf{R}$, recovered 3D \ac{RECIST}-marked lesions $\mathbf{P_H^R}$ and $\mathbf{P_M^R}$, mined lesions $\bigcup_{i=1}^{k-1}\mathbf{P^+_{H,i}}$, and mined hard negatives $\mathbf{P_H^-}$ and $\mathbf{P_M^-}$. 
        The recall numbers are extracted from FROC curves at operation points from FP$=0.125$ to FP$=8$ per volume. Higher recall demonstrates better detection performance.} 
        } \label{exp:add-proposals}
        \begin{tabular}{c|c|cccccc|ccccccc}
        \toprule
            {\bf Exp.} & iter. $k$ & $\mathbf{R}$ & $\mathbf{P_M^R}$ & $\mathbf{P_{M,k-1}^-}$ & $\mathbf{P_H^R}$ & $\bigcup_{i=1}^{k-1}\mathbf{P^+_{H,k}}$ & $\mathbf{P^-_{H,k-1}}$ & 
            \multicolumn{7}{c}{\bf Recall (\%) $@$ FPs$=$[0.125,0.25,0.5,1,2,4,8] per volume} \\
        \midrule
            (a) & 1 & \checkmark &            &            &            &            &            & 12.0 & 21.0 & 31.7 & 45.7 & 53.1 & 61.0 & 66.9 \\
            (b) & 1 & \checkmark & \checkmark &            &            &            &            & 16.0 & 21.7 & 32.9 & 44.3 & 55.7 & 62.6 & 69.5 \\
            (c) & 2 & \checkmark & \checkmark & \checkmark &            &            &            & 19.5 & 27.4 & 36.9 & 48.8 & 55.5 & 64.8 & 70.5 \\
            (d) & 2 & \checkmark & \checkmark & \checkmark & \checkmark &            &            & 17.9 & 31.2 & 45.5 & 52.4 & {\bf 60.5} & {\bf 66.2} & {\bf 72.6} \\
            (e) & 2 & \checkmark & \checkmark & \checkmark & \checkmark & \checkmark &            & 28.9 & 35.5 & 45.5 & 53.1 & 57.6 & 65.2 & 70.7 \\
            (f) & 2 & \checkmark & \checkmark & \checkmark & \checkmark & \checkmark & \checkmark & {\bf 30.4} & 38.6 & {\bf 47.1} & 53.6 & 58.6 & 62.6 & 68.3 \\
            (g) & 3 & \checkmark & \checkmark & \checkmark & \checkmark & \checkmark & \checkmark & 25.5 & {\bf 38.8} & {\bf 47.1} & {\bf 54.3} & 59.5 & 64.1 & 71.7 \\
        \bottomrule
        \end{tabular}
    \end{center}
\end{table*}

\Tab~\ref{exp:add-proposals} presents the performance of the \ac{3DCCN}-based \ac{LPG} when trained with different combinations of harvested lesions. Please note, these results only measure the \ac{LPG} performance, and do not include the effect of the \ac{LPC} filtering. First, as expected, when including the additional labeled proposals, $\mathbf{R_M^U}$, the performance does not improve much over simply using the original \ac{RECIST} marks. This reflects the relatively small size of $\mathbf{R_M^U}$ compared to the entire dataset. However, larger impacts can be seen when including the hard negatives, $\mathbf{P_{M,i}^{-}}$, from the fully-labeled subset. When including hard negatives from our volumes needing harvesting, \ie{},~$\mathbf{P_{H,i}^{-}}$, performance boosts are even greater at the high precision operation points where FPs $\le1$. This validates our \ac{HNSL} approach of using hard-negative cases. Meanwhile, the addition of extra positive samples, $\mathbf{P_{H,i}^{+}}$ and $\mathbf{P_H^R}$, contribute much to the recall when FPs $\ge2$ per volume, as the trained \ac{LPG} becomes much more sensitive. In summary, these results indicate that the harvested lesions and hard negatives can significantly boost how many lesions can be recovered from the DeepLesion dataset.  

Because of the simple one-stage anchor-free architecture, our proposed \ac{3DCCN} processes CT slices at the speed of $26.6$ frames per second (FPS). It runs two times faster than Faster R-CNN~\cite{DBLP:journals/pami/RenHG017}. It largely speeds up the lesion mining pipeline and will satisfy clinical needs better.

\begin{figure}[t!]
    \centering  
    \includegraphics[width=.98\linewidth]{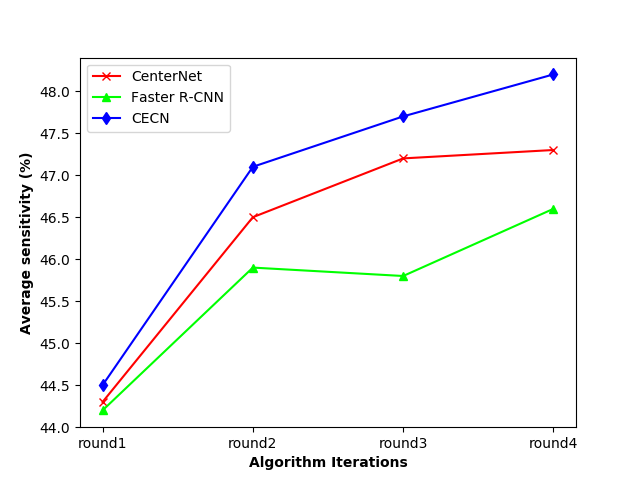}  
    \caption{Lesion-Harvester using different \acp{LPG}, including CenterNet~\cite{DBLP:journals/corr/abs-1904-07850}, Faster R-CNN~\cite{DBLP:journals/pami/RenHG017}, and our proposed \ac{3DCCN}. Mean value of recalls at precision$=[80\%, 85\%, 90\%, 95\%]$ is measured with each ablation study.}  
    \label{fig:Ablation_LPGs}
\end{figure}

In addition to measuring the impact of the different types of harvested lesions, we also ran the Lesion-Harvester pipeline with different \acp{LPG} by replacing \ac{3DCCN} with CenterNet~\cite{DBLP:journals/corr/abs-1904-07850} and Faster R-CNN~\cite{DBLP:journals/pami/RenHG017}. We compare different \acp{LPG} in \Fig~\ref{fig:Ablation_LPGs} and demonstrate that our proposed pipeline generalizes well across   choices of \ac{LPG}. We monitor the mean recall on the validation set and stop algorithm updates after this value has converged. We plot the mean recalls of the first \textcolor{black}{four} iterations in \Fig~\ref{fig:Ablation_LPGs}. Compared with the baseline $36.7\%$ recall provided by \ac{RECIST} marks, Lesion-Harvesters using Faster R-CNN~\cite{DBLP:journals/pami/RenHG017}, CenterNet~\cite{DBLP:journals/corr/abs-1904-07850}, and our proposed \ac{3DCCN} have achieved $9.9\%$, $10.6\%$, and $11.5\%$ improvements, respectively. In spite of the used \ac{LPG}, our proposed pipeline can roughly recover $10\%$ of unlabeled lesions. We also  note that more powerful \acp{LPG} could result in faster convergence rates as well as better recovery rates.

\subsection{Ablation Study: \ac{LPC}}
\begin{table}[t!]
    \centering  
    \caption{Mean recalls (\%) from \ac{R@80P} to \ac{R@95P} of different lesion proposal classifiers.\label{tab:lpc-gl}}
    \begin{tabular}{lccccc}  
        \toprule  
        \textbf{Method} & R@ & R@ & R@ & R@ & R@ \\  
        & 80P & 85P & 90P & 95P & Avg. \\
        \midrule  
        \acs{MLC3D}~\cite{DBLP:journals/tbe/DouCYQH17} & 42.4          & 41.2          & 39.8          & 38.0          & 40.3 \\  
        ResNet-3D                                      & 44.8          & 43.3          & 42.1          & 40.0          & 42.6 \\  
        \acs{GL3D}                                     & 46.0          & 43.4          & 42.6          & \textbf{40.7} & 43.2 \\  
        ResNet-MV                                      & \textbf{47.9} & 46.2          & 43.1          & 40.1          & 44.3 \\  
        \acs{MVCNN}                                    & 47.4          & \textbf{46.3} & \textbf{44.7} & 40.4          & \textbf{44.7} \\  
        \bottomrule
    \end{tabular}
\end{table}

We validate our choice of \ac{LPC} by first comparing the performance of different global-local classifiers evaluated on at the first iteration of our method. We compare our multi-view \ac{MVCNN} with two alternatives: a 2D variant that only accepts the axial view as input and a 3D version of ResNet-18~\cite{DBLP:conf/cvpr/HaraKS18}. We measure results when using the raw \ac{LPG} detection score ($s_{g}$), \ac{LPC} classification probability ($s_{c}$), and the final lesion score ($s_{\{p|g,c\}}$) of \Eq~\ref{eqn:proposal_score}. As can be seen in \Fig~\ref{fig:exp-lpc}, not all \acp{LPC} outperform the raw detection scores ($s_{g}$). However, they all benefit from the re-scoring of \Eq~\ref{eqn:proposal_score} using detection scores. Out of all options, the multi-view approach works the best. In addition to its high performance, it also has the virtue of being much simpler and faster than a full 3D approach. 

In addition, we also validate the global-local feature. To do this, we compare against ResNet classifiers, both multi-view and 3D, that forego the global-local feature concatenation. We also test against a recent multi-scale approach called \ac{MLC3D}~\cite{DBLP:journals/tbe/DouCYQH17}, which is designed for lesion \ac{FP} filtering. We also implemented the 3D classifier presented in~\cite{ieee/miccai/DingLHW17}; however, it did not converge with DeepLesion as it uses a much shallower network than ResNet18. For all, we evaluated using different input sizes centered at the lesion: (24$\times$24$\times$4), (48$\times$48$\times$8), (64$\times$64$\times$16), and (128$\times$128$\times$32). Here we only report results using the best scale for each, but our supplementary material contains more complete results. As can be seen in \Tab~\ref{tab:lpc-gl}, \ac{MLC3D}~\cite{DBLP:journals/tbe/DouCYQH17} does not deliver any improvement over standard ResNet-3D. In contrast the global-local variants outperform their standard counterparts demonstrating the effectiveness of our multi-scale approach design. Again, the \ac{MVCNN} delivers the best classification performance.

\subsection{Main Result: Detectors Trained on Harvested Lesions}
While the above demonstrated that we can successfully harvest missing lesions with high precision, it remains to be demonstrated how beneficial this is. To this end, we train state-of-the-art detection systems with and without our harvested lesions and also compare against some alternative approaches to manage missing labels. 

\begin{table*}[t!]
    \centering
    \caption{Evaluation of detectors trained with and without mined lesions on $\mathbf{V_D^{test}}$.} \label{tab:detection3d} 
    \begin{tabular}{l|c|c|ccc|ccccccc|c}
        \toprule 
        {\bf Method} & {\bf Backbone} & {\bf Input} & $R$ & $\mathbf{P_H^+}$ & $\mathbf{P^-}$
        & \multicolumn{7}{c|}{\bf Recall (\%) $@$ FPs$=$[0.125,0.25,0.5,1,2,4,8] per volume} & {\bf AP} \\ 
        \midrule 
        MULAN~\cite{ieee/miccai/YanTPSBLS19}  & 2.5D DenseNet-121 & 9 & \checkmark & &             & 11.43 & 18.69 & 26.98 & 38.99 & 50.15 & 60.38 & 69.71 & 41.8 \\
        \midrule
        Faster R-CNN~\cite{DBLP:journals/pami/RenHG017}        & 2.5D DenseNet-121 & 9 & \checkmark & &              & 07.20 & 13.21 & 20.97 & 31.27 & 43.87 & 54.92 & 64.20 & 34.2 \\
        ~ w/ OBHS~\cite{DBLP:journals/pami/RenHG017} & 2.5D DenseNet-121 & 9 & \checkmark &            &             & 02.51 & 04.77 & 08.40 & 17.46 & 22.65 & 34.24 & 47.01 & 17.3 \\
        ~ w/ OBSS~\cite{DBLP:conf/bmvc/WuBSNCD19}  & 2.5D DenseNet-121 & 9 & \checkmark & &         & 06.85 & 12.58 & 21.92 & 32.39 & 44.53 & 57.08 & 67.91 & 36.3 \\
        ~ w/ M-OBHS & 2.5D DenseNet-121 & 9 & \checkmark &            &                                      & 08.53 & 13.67 & 22.89 & 34.46 & 46.85 & 58.31 & 68.08 & 38.3 \\
        Faster R-CNN & 2.5D DenseNet-121 & 9 & \checkmark & \checkmark &                                     & 10.35 & 15.98 & 25.02 & 35.63 & 47.15 & 57.74 & 66.77 & 38.5 \\
        ~ w/ ULDor~\cite{DBLP:conf/isbi/TangYTLXS19} & 2.5D DenseNet-121 & 9 & \checkmark & \checkmark & \checkmark     
        & {\bf 22.95} & {\bf 30.15} & {\bf 38.09} & {\bf 46.90} & 55.23 & 63.05 & 70.48 & 51.3 \\
        \midrule 
        CenterNet & DenseNet-121 & 3 & \checkmark &            &                                        & 12.91 & 19.89 & 26.19 & 35.77 & 45.32 & 56.94 & 67.83 & 41.0 \\
        CenterNet & DenseNet-121 & 3 & \checkmark & \cite{DBLP:conf/prcv/WangLZZH19} & & 12.47 & 18.12 & 25.36 & 35.03 & 46.52 & 58.20 & 68.95 & 41.3 \\
        CenterNet & DenseNet-121 & 3 & \checkmark & \checkmark &                                        & 14.38 & 19.75 & 28.04 & 36.89 & 46.82 & 58.94 & 68.70 & 42.8 \\
        CenterNet w/ \ac{HNSL} & DenseNet-121 & 3 & \checkmark & \checkmark & \checkmark                & 19.26 & 25.87 & 34.54 & 43.17 & 53.34 & 63.08 & 71.68 & 48.3 \\
        \midrule 
        \ac{3DCCN} & 2.5D DenseNet-121 & 9 & \checkmark &            &                                         & 11.92 & 18.42 & 27.54 & 38.91 & 50.15 & 60.76 & 69.82 & 43.0 \\
        \textcolor{black}{\ac{3DCCN}} & 2.5D DenseNet-121 & 9 & \checkmark & \cite{DBLP:conf/prcv/WangLZZH19}&   & 15.88 & 20.98 & 29.40 & 38.36 & 47.89 & 58.14 & 67.89 & 43.2 \\
        \ac{3DCCN} & 2.5D DenseNet-121 & 9 & \checkmark & \checkmark &                                         & 13.40 & 19.16 & 27.34 & 37.54 & 49.33 & 60.52 & 70.18 & 43.6 \\
        \ac{3DCCN} w/ \ac{HNSL} & 2.5D DenseNet-121 & 9 & \checkmark & \checkmark & \checkmark& 19.86 & 27.11 & 36.21 & 46.82 & {\bf 56.89} & {\bf 66.82} & {\bf 74.73} & {\bf 51.9} \\
        \bottomrule
    \end{tabular}
\end{table*}

\subsubsection{Using Harvested Lesions and Hard Negative Examples}
After our method converged, we fused mined lesions and hard negatives, \ie{}, $\mathbf{P^+_H}=\bigcup_{i=1}^{3}\mathbf{P^+_{H,i}}$ and $\mathbf{P^-}=\mathbf{P_{H,3}^-}\cup\mathbf{P_{M,3}^-}$, respectively. We tested CenterNet~\cite{DBLP:journals/corr/abs-1904-07850}, Faster R-CNN~\cite{DBLP:journals/pami/HeGDG20}, and our \ac{3DCCN} (used now as a detector instead of an \ac{LPG}), trained both on the original DeepLesion \ac{RECIST} marks and on the data augmented with our harvested lesions. Since \ac{MULAN} requires tags, which are not available for harvested lesions, we only test it using the publicly-released model. As well, we also test the impact of our hard negatives ($\mathbf{L^-}$), with our proposed \ac{HNSL} on CenterNet and \ac{3DCCN}. We do not test the \ac{HNSL} with Faster R-CNN, since it is not compatible with two-stage anchor-based systems. \textcolor{black}{Instead, we follow Tang \etal{},'s (ULDor)~\cite{DBLP:conf/isbi/TangYTLXS19}, which is a hard-negative approach designed for Faster R-CNN that defines an additional non-lesion class containing all hard negative examples.} We test all detector variants on the unseen fully labeled $\mathbf{V_D^{test}}$ data. 

As \Tab~\ref{tab:detection3d} demonstrates, using the harvested lesions to train detectors can provide significant boosts in recall and precision for all methods. For instance, the extra mined lesions $\mathbf{P_H^+}$ boosts Faster R-CNN's detection performance by $4.3\%$ in \acl{AP}. When incorporating hard negatives using the \ac{HNSL}, CenterNet and \ac{3DCCN} both benefited even more, with additional boosts of $5-8\%$ \ac{AP}. In total, the harvested prospective positive and hard negative lesions are responsible for a boost of $7.2\%$ and $8.9\%$ in \ac{AP}, for CenterNet and \ac{3DCCN}, respectively, representing a dramatic boost in performance. \textcolor{black}{Despite the great differences in architecture, incorporating harvested hard negatives as an extra non-lesion class~\cite{DBLP:conf/isbi/TangYTLXS19} also significantly boosts Faster R-CNN performance, further demonstrating the broad impact of the Lesion-Harvester proposals.} Finally, we also note that our \ac{3DCCN} outperforms the state-of-the-art lesion detection model, \ac{MULAN}~\cite{ieee/miccai/YanTPSBLS19}, even when no mined lesions are used. The addition of mined lesions, which is not directly applicable with \ac{MULAN}, further boosts the performance gap to $10.1\%$. This further validates our \ac{LPG} design choices and represents an additional contribution of this work, in addition to our main focus of lesion harvesting.

\subsubsection{Alternative Missing Label Approaches}
We also evaluated other strategies for managing missing labels. These include \acf{HNC}~\cite{DBLP:journals/pami/RenHG017} and \acf{OBSS}~\cite{DBLP:conf/bmvc/WuBSNCD19}, both of which are designed only for two-stage anchor-based detection networks, meaning they are incompatible with our one-stage \ac{3DCCN}. Thus, we use Faster R-CNN as baseline and compared \ac{HNC} and \ac{OBSS} to training using our harvested prospective positive lesions, $\mathbf{P_H^+}$. \ac{HNC} only uses proposals with small overlap to a true positive as negative examples for the second stage classifier. As \Tab~\ref{tab:detection3d} demonstrates, the baseline Faster R-CNN performed with $34.2\%$ \ac{AP}, whereas using \ac{HNC} reduced the \ac{AP} to $17.3\%$, demonstrating that simply ignoring the predominant background negatives causes large performance degradation. Thus, we also re-trained Faster R-CNN with a modified \ac{HNC} (M-OBHS), which preserves all background samples but raises weights of overlapping proposals to be twice as much as positive and standard background cases. This variant achieved $38.3\%$ \ac{AP}. As a different strategy, \ac{OBSS} reduces the contributions of proposals which have small overlaps with the ground truth boxes but keeps all background cases. This strategy increased the \ac{AP} to $36.3\%$. Finally, our method that trains Faster R-CNN with harvested prospective positive lesions achieved the best performance at $38.5\%$ \ac{AP}, with markedly higher recalls at lower tolerated \acp{FP}. This demonstrates that completing the label set with harvested lesions $\mathbf{P_H^+}$ can provide greater boosts in performance than these alternative strategies. 

Moving on to a more general approach, we also compare against the \ac{MGTM}-\ac{SLLP}~\cite{DBLP:conf/prcv/WangLZZH19} lesion mining algorithm. As shown in \Tab~\ref{tab:detection3d}, \ac{MGTM}-\ac{SLLP} boosts the \ac{AP} of CenterNet from $41.0\%$ to $41.3\%$, while training with $\mathbf{P_H^+}$ results in a much larger boost to $42.8\%$. Similarly, training with $\mathbf{P_H^+}$ garners higher boosts in \ac{AP} when using  our \ac{3DCCN} detector. Unlike \ac{MGTM}-\ac{SLLP}, Lesion-Harvester explores the whole \ac{CT} volume, iteratively improving performance after each round, and integrates hard negatives within the training of \ac{LPG}. We surmise these differences explain the increased performance even when comparisons are limited to only using the harvested positive proposals of $\mathbf{P_H^+}$. Finally, as we demonstrate, when using our harvested hard-negative proposals performance can be increased even further.

\section{Conclusions} \label{sec:conclusions}
We present an effective framework to harvest lesions from incompletely labeled datasets. Leveraging a very small subset of fully-labeled data, we chain together an \ac{LPG} and \ac{LPC} to iteratively discover and harvest unlabeled lesions. We test our system on the DeepLesion dataset  and show that after only annotating $~5\%$ of the volumes we can successfully harvest $9,805$ additional lesions, which corresponds to $47.9\%$ recall at $90\%$ precision, which is a boost of $11.2\%$ in recall over the original \ac{RECIST} marks. Since, our proposed method is an open framework, it can accept any state-of-the-art \ac{LPG} and \ac{LPC}, allowing it to benefit from future improvements in these two domains.

Our work's impact has several facets. For one, in terms of DeepLesion specifically, the lesions we harvest and publicly release enhance the utility of an already invaluable dataset. As we demonstrated, training off-the-shelf detectors on our harvested lesions allows them to outperform the current best performance on the DeepLesion dataset by margins as high as $10\%$ \ac{AP}, which is a significant boost in performance. Furthermore, we expect our harvested lesions will prove useful to many applications beyond detection, \eg{},~radionomics studies. More broadly, our results indicate that the lesion harvesting framework is a powerful means to complete \ac{PACS}-derived datasets, which we anticipate will be an increasingly important topic. Thus, this approach may help further expand the scale of data for the medical imaging analysis field. 

Important contributions also include our proposed \ac{3DCCN} \ac{LPG}, which outperforms the current state-of-the-art \ac{MULAN} detector and helps push forward the lesion detection topic.  In addition, the introduced \ac{P3D} IoU metric acts as a much better evaluation metric for detection performance than current practices. As such, the adoption of the \ac{P3D} metric as a standard for DeepLesion evaluation should better rank methods going forward. \textcolor{black}{Future work should include investigating different \ac{LPG} paradigms, \eg{}, full 3D approaches, executing user studies, \eg{}, to assess and compare with clinician performance, and exploring active learning to more efficiently choose which volumes to label. In addition, it is also crucial to measure the impact of differing labor budgets on harvesting performance. Finally, we estimate that possibly $40\%$ of the lesions still remain unlabeled in DeepLesion. It is possible to calibrate our proposed \ac{LPG} to detect lesions with above $90\%$ recall, however, the challenge is to separate true lesions from thousands of false positives. Therefore, one other promising avenue is to model the relationship between lesions proposals for false positive reduction, which may lead to non-parametric matching and graph-based connections between instances within the dataset. We hope the public annotations and benchmarks we release will spur further solutions to this important problem.}

\section*{Supplementary Material}  

\subsection{Evaluation of Generated 3D Boxes}  
\begin{figure}[t!]
  \centering 
  \includegraphics[trim=0.2in 0 0.2in 0,clip,width=\linewidth]{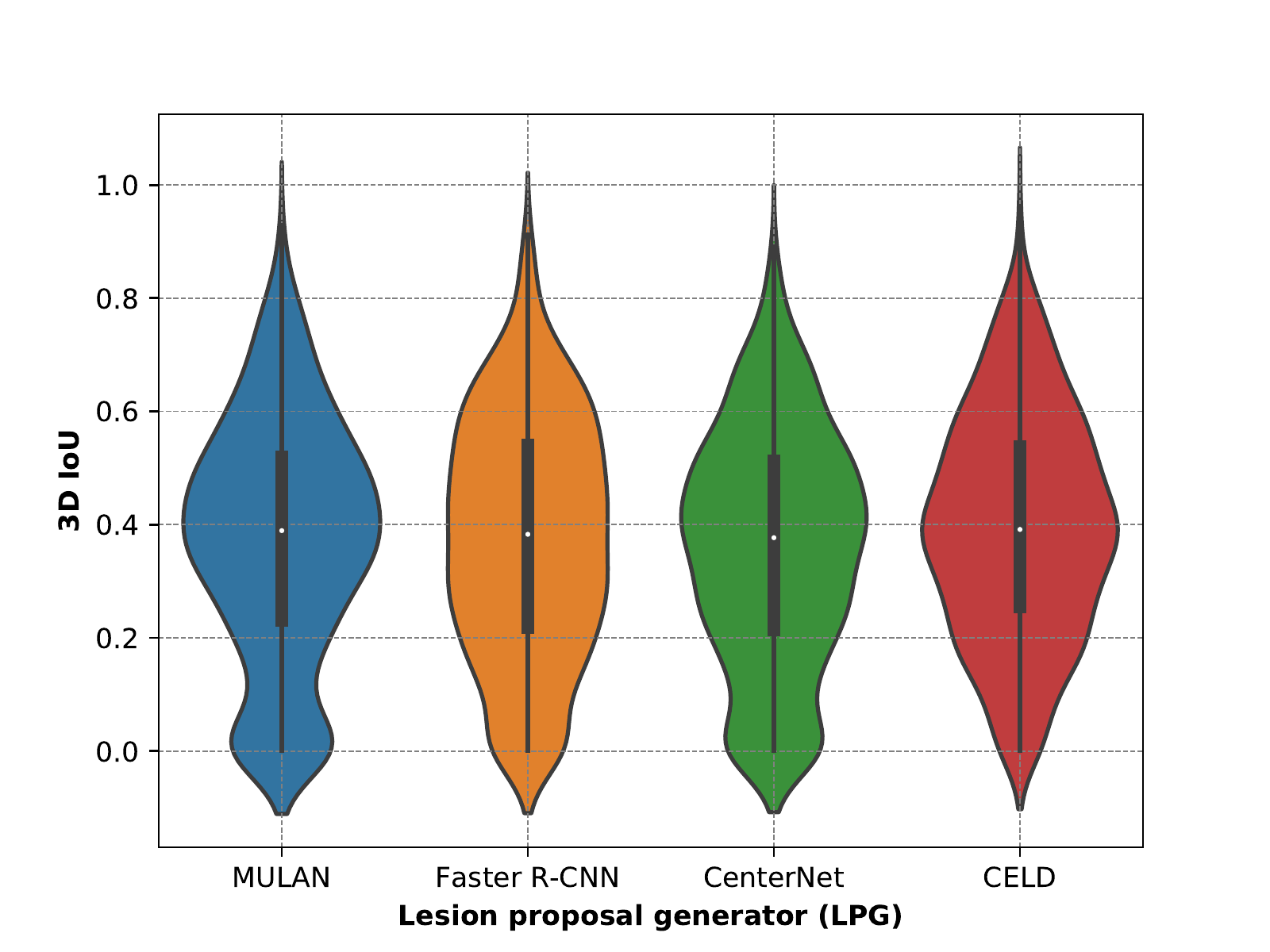}
  \caption{\textbf{Violin plot.} Shown is the distribution of 3D \acp{IoU} of generated 3D proposals evaluated on the testing sub-set with full 3D annotations. Detection models including \ac{MULAN}~\cite{ieee/miccai/YanTPSBLS19}, Faster R-CNN~\cite{DBLP:journals/pami/RenHG017}, CenterNet~\cite{DBLP:journals/corr/abs-1904-07850}, and our proposed \ac{3DCCN} are evaluated}  
  \label{fig:3DBox-Evaluation}
\end{figure}

We evaluate the generated 3D boxes using the $272$ testing cases with full 3D box annotations. With each annotated lesion, we compare the generated 3D box with the ground-truth 3D box with 3D-\ac{IoU}. To demonstrate the quality of reconstructed 3D boxes, we show violin plot of 3D-\acs{IoU} in \Fig~\ref{fig:3DBox-Evaluation}. Regardless of the chosen \ac{LPG}, medians are around $0.4$ 3D-\ac{IoU}, which is considered as a high overlap value in 3D detection. All of the first quartiles are above $0.2$ 3D-\ac{IoU} showing that the most of the reconstructed 3D boxes are of high qualities.  

\subsection{Lesion Type Analysis of Mined Lesions}
\begin{figure}[t!]  
  \centering  
  \includegraphics[width=\linewidth]{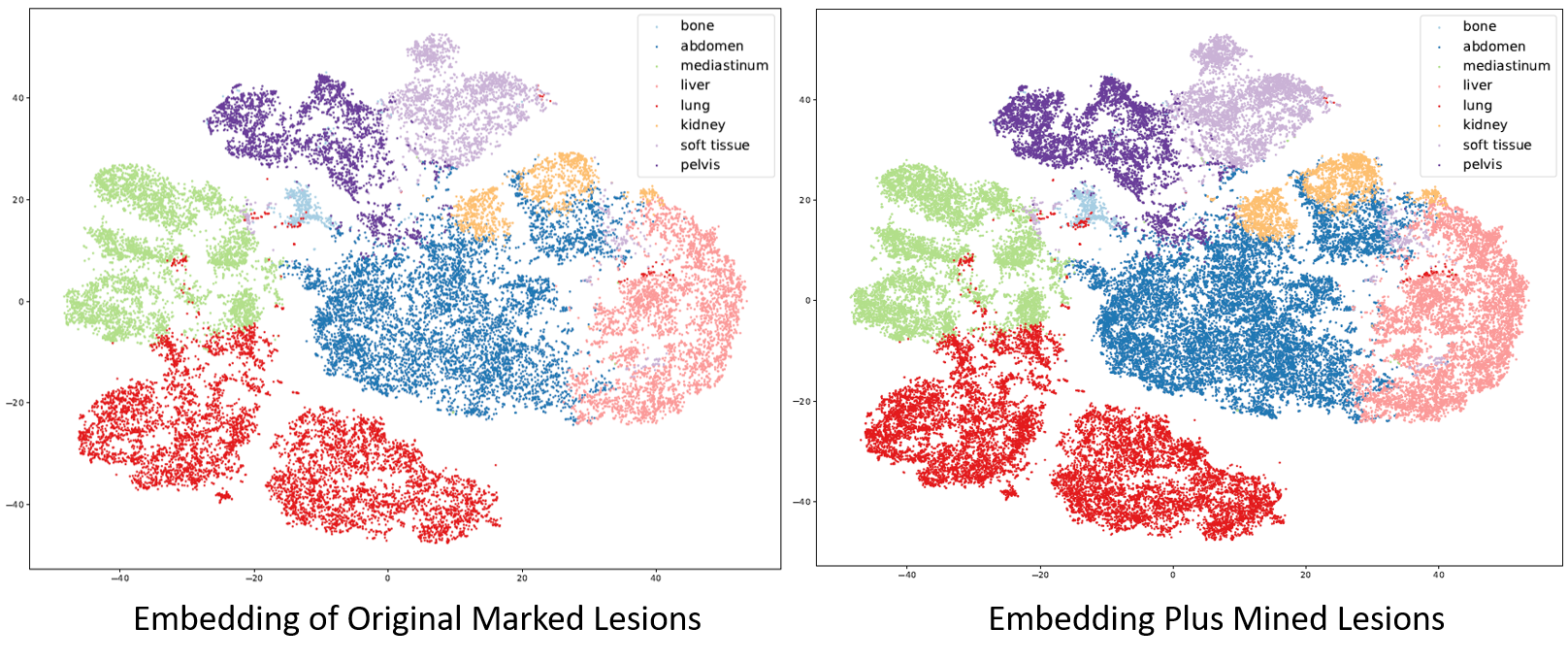}  
  \caption{\textcolor{black}{Embedding of the original \ac{RECIST}-marked lesions and plus mined lesions.}}
  \label{fig:review-example-2}
\end{figure}

DeepLesion contains various types of lesions, but the major categories are lung, liver, kidney lesions as well as enlarged lymph nodes. Therefore, the proposed algorithm is more sensitive to these major lesion categories. We applied the lesion embedding and classification method proposed by Yan \etal{},~\cite{DBLP:conf/cvpr/YanWLZHBS18} to our mined lesions. Visualization of the embeddings of the original \ac{RECIST}-marked lesions and mined lesions are show in \Fig~\ref{fig:review-example-2}. We observe that the mined lesions roughly follow the distribution of the original lesion distribution.

\subsection{Comparison of \acp{LPC}}  
\begin{figure}[t!]  
  \centering  
  \includegraphics[width=\linewidth]{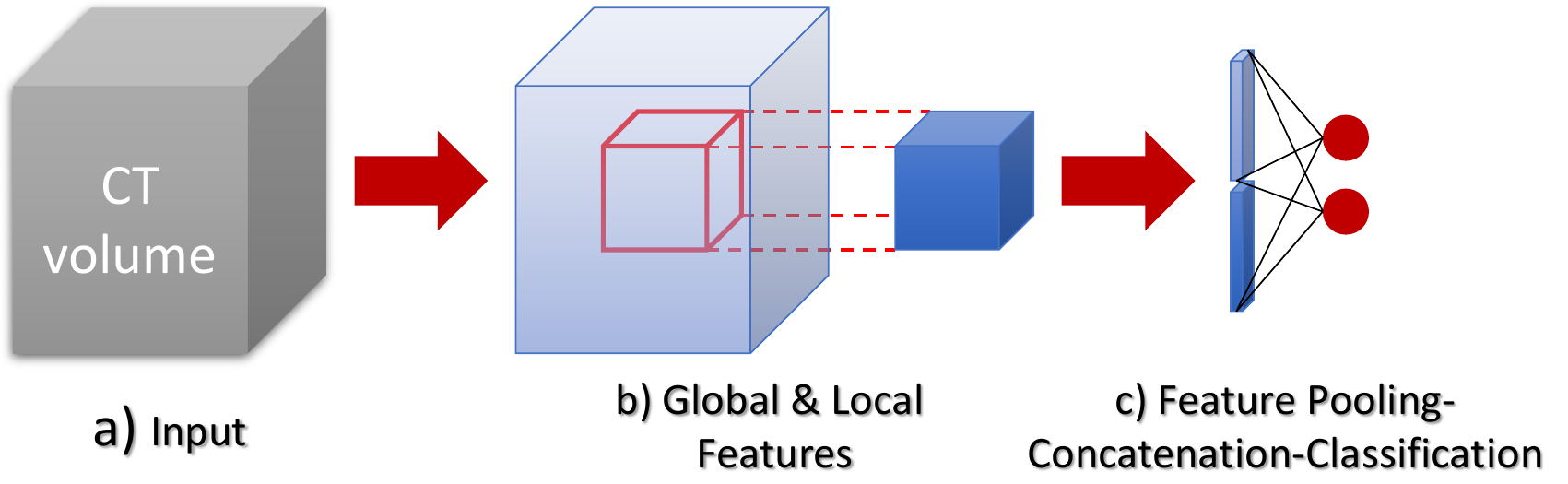}  
  \caption{\Acf{GL3D}.} \label{fig:other-lpcs}
\end{figure}

To demonstrate the effectiveness of our proposed global-local design, we evaluated different input scales including scale-1 (24$\times$24$\times$4), scale-2 (48$\times$48$\times$8), scale-3 (64$\times$64$\times$16), and scale-4 (128$\times$128$\times$32) and each cropped sub-volume is centered at the centroid of the detected lesion proposal. The four scales cover $57.8\%$, $91.8\%$, $97.5\%$, and $99\%$ \ac{RECIST} markers in $\{\mathbf{R_H}\cup\mathbf{R_M}\}$, respectively. With each scale, we train a ResNet-3D~\cite{DBLP:conf/cvpr/HaraKS18} and measure the mean of recalls from \ac{R@80P} to \ac{R@95P}. From \Tab~\ref{tab:lpc-gl-full}, we observe that the mean recall decreases along with the input scale increases. It indicates direct mixing of foreground and background should harm lesion recognition in DeepLesion. \ac{MLC3D}~\cite{DBLP:journals/tbe/DouCYQH17} combines predictions of scale-1, scale-2, and scale-3, however, it does not deliver any improvement. We then test our global-local design using ResNet-3D~\cite{DBLP:conf/cvpr/HaraKS18}. \Fig~\ref{fig:other-lpcs} shows our design of \ac{GL3D}, which utilizes lesion proposals to extract local features for classification. In \Tab~\ref{tab:lpc-gl-full}, \ac{GL3D} outperforms its counterpart ``ResNet-3D with scale-4'' by $2.9\%$, demonstrating the effectiveness of modeling foreground and background separately. The ResNet-3D with ``box'' input processed only the local (or foreground) image. Compared with \ac{GL3D}, it has higher \acs{R@95P} value indicating better specificity on certain lesions, however, its lower \ac{R@80P}, \ac{R@85P}, and \ac{R@90P} values showing worse generalizability since no background context has been considered. Finally, the \ac{MVCNN} combines multi-view structure with the global-local design and delivers the best classification performance in DeepLesion.  

\begin{table}[t!]
  \centering  
  \caption{Comparison of \acfp{LPC}. Detection recalls (R) at precisions (P) from $80\%$ to $95\%$ are reported.}
  \label{tab:lpc-gl-full}
  \begin{tabular}{lcccccc}  
  \toprule
  \textbf{Method} & \textbf{Input} & R@ & R@ & R@ & R@ & R@ \\  
  & & 80P & 85P & 90P & 95P & Avg. \\
  \midrule
  ResNet-3D & scale-1               & 44.8 & 43.3 & 42.1 & 40.0 & 42.6 \\  
  ResNet-3D & scale-2               & 42.4 & 43.1 & 41.3 & 38.8 & 41.0 \\  
  ResNet-3D & scale-3               & 41.9 & 41.7 & 40.7 & 38.7 & 40.7 \\  
  ResNet-3D & scale-4               & 41.2 & 40.7 & 39.2 & 38.3 & 40.3 \\  
  \acs{MLC3D}~\cite{DBLP:journals/tbe/DouCYQH17} & scale-1\&2\&3 & 42.4 & 41.2 & 39.8 & 38.0 & 40.3 \\  
  ResNet-3D   & scale-4, box        & 45.7 & 43.2 & 42.3 & \textbf{41.2} & 43.1 \\  
  \acs{GL3D}  & scale-4, box        & 46.0 & 43.4 & 42.6 & 40.7 & 43.2 \\  
  ResNet-MV   & scale-4, box        & \textbf{47.9} & 46.2 & 43.1 & 40.1 & 44.3 \\  
  \acs{MVCNN} & scale-4, box        & 47.4 & \textbf{46.3} & \textbf{44.7} & 40.4 & \textbf{44.7} \\  
  \bottomrule
  \end{tabular}
\end{table}

\bibliographystyle{IEEEtran}
\bibliography{IEEEabrv,egbib}  

\end{document}